%% file: neurips_2024.tex
\documentclass{article}

\PassOptionsToPackage{numbers, compress}{natbib}

\usepackage[final]{neurips_2024}

\usepackage[utf8]{inputenc} 
\usepackage[T1]{fontenc}    
\usepackage{hyperref}       
\usepackage{url}            
\usepackage{booktabs}       
\usepackage{amsfonts}       
\usepackage{nicefrac}       
\usepackage{microtype}      
\usepackage{xcolor}         
\usepackage{natbib}
\usepackage{algorithm}
\usepackage{algpseudocode}
\usepackage{amsmath}
\usepackage{bbm}
\usepackage{caption}
\usepackage{subfigure}
\usepackage{subcaption}
\usepackage{comment}
\usepackage{multirow}
\usepackage{bbding}
\usepackage{graphicx}
\usepackage{enumitem}

\title{MutaPLM: Protein Language Modeling for Mutation Explanation and Engineering}

\author{%
Yizhen Luo$^{1,2,}$\thanks{Equal contribution}\ \,, Zikun Nie$^{1,2,\dag}$, Massimo Hong$^{1,2}$,\\ \textbf{Suyuan Zhao$^{1,2}$, Hao Zhou$^{1,}$\thanks{Corresponding authors}\ \,, Zaiqing Nie$^{1,3,*}$}\\
  $^{1}$Institute of AI Industry Research (AIR), Tsinghua University\\
  $^{2}$Department of Computer Science and Technology, Tsinghua University\\
  $^{3}$Pharmolix Inc.\\
  \texttt{\{yz-luo22,nzk24,hongcd21,zhaosy23\}@mails.tsinghua.edu.cn}\\
  \texttt{\{zhouhao,zaiqing\}@air.tsinghua.edu.cn}     \\
}

\begin{document}

\maketitle

\begin{abstract}
Studying protein mutations within amino acid sequences holds tremendous significance in life sciences. Protein language models (PLMs) have demonstrated strong capabilities in broad biological applications. However, due to architectural design and lack of supervision, PLMs model mutations implicitly with evolutionary plausibility, which is not satisfactory to serve as explainable and engineerable tools in real-world studies. To address these issues, we present MutaPLM, a unified framework for interpreting and navigating protein mutations with protein language models. MutaPLM introduces a protein \textit{delta} network that captures explicit protein mutation representations within a unified feature space, and a transfer learning pipeline with a chain-of-thought (CoT) strategy to harvest protein mutation knowledge from biomedical texts. We also construct MutaDescribe, the first large-scale protein mutation dataset with rich textual annotations, which provides cross-modal supervision signals. Through comprehensive experiments, we demonstrate that MutaPLM excels at providing human-understandable explanations for mutational effects and prioritizing novel mutations with desirable properties. Our code, model, and data are open-sourced at \url{https://github.com/PharMolix/MutaPLM}.
\end{abstract}

\input{1_introduction}

\input{2_related_work}
\input{3_method}

\input{4_experiments}
\input{5_conclusion}

\begin{ack}
This research is supported by the National Key R\&D Program of China (No. 2022YFF1203002) and PharMolix Inc.
\end{ack}

\bibliography{refs.bib}

\appendix

\input{6_appendix}


\newpage

\input{7_checklist}

\end{document}

%% file: 1_introduction.tex
\section{Introduction}




Studying protein evolution through mutations within amino acid sequences is a central research topic in life sciences \cite{pal2006integrated, soskine2010mutational, reva2011predicting}. Despite immense research efforts, a large number of protein mutations with biological significance remain under-explored, highlighting the demand for in-silico tools to model these mutations. Practically, the tool should meet two requirements. First, it should be \textbf{explainable}, providing insightful and human-understandable interpretations for mutational effects. This is crucial for broad biological applications ranging from identifying immune-escape pathogens \cite{harvey2021sars, hu2022increased} to interpreting the mechanisms of human diseases \cite{thomas1995defective, dobson2001structural}. Additionally, the tool should be \textbf{engineerable}, proposing protein mutations that satisfy desirable properties such as catalytic activity and thermostability. This process is known as directed evolution \cite{turner2009directed, arnold2018directed}, the most prevailing approach for protein design in the laboratory, which offers substantial benefits across various application fields, including industry \cite{arnold1999directed}, biotechnology \cite{sawano2000directed}, and therapeutics \cite{boder2000directed}.


To achieve these goals, deep learning models \cite{sun2017sequence, gao2020deep, bepler2021learning} have emerged to capture evolutionary information from protein sequences. Recently, the development of protein language models (PLMs) \cite{madani2020progen, elnaggar2021prottrans, ferruz2022protgpt2, lin2023evolutionary, truong2024poet} has brought a paradigm shift in computational biology. By self-supervised learning \cite{liu2021self} on evolutionary-scale databases \cite{suzek2015uniref, steinegger2018clustering}, PLMs have achieved great success in various biological applications, including structure prediction \cite{lin2023evolutionary, weissenow2022protein} and protein design \cite{ferruz2022protgpt2, ferruz2022controllable}. Additionally, PLMs have demonstrated zero-shot capabilities in predicting and optimizing evolutionary plausibility \cite{meier2021language, hsu2022learning, zhao2024contrastive}, a continuous value indicating whether a mutation is favored by natural selection.

Despite their promising advancements, we argue that existing PLMs are not yet satisfactory as explainable and engineerable tools for handling protein mutations. Regarding mutation explanation, PLMs' implicit interpretation with evolutionary plausibility is overly vague, lacking detailed information for mutational effects such as specific alterations in protein functions and impacts on organisms. Regarding mutation engineering, PLMs can only propose evolutionary-plausible mutations, which may be misaligned with human preferences in real-world practices of directed evolution. For example, enhancing the catalytic activity of an enzyme from a bacterium could be detrimental to its survival due to increased energy costs but beneficial for industrial applications. In such scenarios, the utility of PLMs in assisting protein engineering is significantly compromised.


In this paper, we aim to develop explainable and engineerable PLMs by explicitly modeling protein mutations. However, conventional PLMs based on the Transformers \cite{vaswani2017attention} architecture provide context-aware representations for each amino acid, which are inadequate for capturing the discrepancies between the wild-type and its mutant within a unified feature space. Besides, there is a lack of supervision signals necessary for comprehending the intricate impacts of protein mutations, which require extensive background knowledge, including protein structures, protein functions, and mechanisms of biological processes.



To address these issues, we envision that (1) mutation representations could be captured from the variations of PLM representations between the wild-type and its mutant with appropriate architecture, and (2) expert-written texts from protein databases and biomedical publications provide rich cross-modal supervision for learning protein mutations. Specifically, we propose \textbf{MutaPLM}, a unified framework for interpreting and navigating \textbf{Muta}tions with \textbf{P}rotein \textbf{L}anguage \textbf{M}odels. We introduce a protein \textit{delta} network that translates between mutations and protein \textit{delta} features, formulating a unified feature space aligned with textual semantics. We develop a transfer learning pipeline with a chain-of-thought (CoT) strategy \cite{wei2022chain} to harvest protein mutation knowledge from biomedical texts. Additionally, we construct MutaDescribe, the first large-scale dataset containing diverse protein mutations and rich textual annotations of their effects. Using natural language as a friendly interface, the dataset facilitates the training and evaluation of mutation explanation and engineering.

Through comprehensive experiments, we demonstrate that MutaPLM is a versatile, explainable, and engineerable tool for assisting protein mutation studies. In mutation explanation, MutaPLM outperforms the strongest baseline model by 6.5\% in ROUGE-L, and 19.4\% of the predicted mutational effects are regarded as accurate and insightful by human experts. In mutation engineering, our model achieves an average of 0.409 recall scores on top-50 mutation proposals navigated by free-text instructions, improving ESM-2 \cite{lin2023evolutionary} by 1.6-fold.

Our contributions are summarized as follows:

\begin{itemize}[noitemsep,topsep=0pt,parsep=5pt,partopsep=0pt,leftmargin=20pt]
\item We propose MutaPLM, a unified framework that enables protein language models to capture mutations explicitly using a protein \textit{delta} network and cross-modal supervision. 
\item We build MutaDescribe, the first dataset with detailed textual annotations for protein mutations.
\item We validate the effectiveness of MutaPLM in explaining and engineering protein mutations through comprehensive experiments.
\end{itemize}

%% file: 2_related_work.tex
\section{Related Work}
\subsection{Protein Language Models}
In analogy to large language models (LLMs) \cite{touvron2023llama, jiang2023mistral, achiam2023gpt, bi2024deepseek} in natural language processing (NLP), protein language models (PLMs) such as ProteinBERT \cite{brandes2022proteinbert}, ProtTrans \cite{elnaggar2021prottrans}, ProtGPT2 \cite{ferruz2022protgpt2}, and ESM series \cite{rives2021biological, lin2023evolutionary, hayes2024simulating} have surged in modeling protein sequences. Pre-trained by masked language modeling \cite{devlin2019bert} or auto-regressive language modeling \cite{radford2018improving} on evolutionary-scale protein databases, PLMs have demonstrated outstanding predictive power on protein secondary and tertiary structures \cite{weissenow2022protein}, protein functions \cite{unsal2022learning} and protein-protein interactions \cite{hu2024improving}. More recently, explorations on PLMs unifying protein sequences and natural language \cite{xu2023protst, luo2024toward, zhuo2024protllm, lv2024prollama} have attracted rising research interest, as texts provide unstructured knowledge and a friendly user interface for studying proteins. Notably, a contemporary work \cite{yin2024multi} proposes to perform text-based protein editing by directly generating the mutated protein sequence. Unfortunately, none of the existing PLMs qualifies as an explainable and engineerable tool in modeling protein mutations, mainly owing to architectural design and lack of supervision.


\subsection{Protein Mutation Modeling}
Previous works formulate mutation explanation as learning the 'local fitness landscape', a mapping from protein sequences to specific functional activity scores \cite{romero2009exploring}. Models for protein fitness prediction could be categorized as (1) alignment-based models \cite{hopf2017mutation, laine2019gemme} trained on multiple sequence alignments (MSAs) \cite{jeanmougin1998multiple}, (2) PLM models \cite{ferruz2022protgpt2, lin2023evolutionary} trained on large-scale unaligned sequences, (3) inverse-folding models \cite{hsu2022learning, dauparas2022robust} that learn protein fitness through structure-conditioned sequence distributions, and (4) hybrid models \cite{rao2021msa, notin2022trancepteve} that combine both PLMs and MSAs. The evaluations are performed as per wild-type protein on deep mutation scanning (DMS) \cite{fowler2014deep} or clinical variant \cite{landrum2018clinvar} benchmarks. In this work, we formulate mutation explanation as a more challenging task that aims at providing textual descriptions of mutational effects for arbitrary wild-type protein and mutation.


The traditional mutation engineering \cite{turner2009directed, arnold2018directed} task aims at generating protein mutants with high fitness scores. One line of work leverages generative models including variational auto-encoders (VAEs) \cite{brookes2019conditioning}, generative language models \cite{stanton2022accelerating} and diffusion models \cite{gruver2024protein} to directly generate the protein sequence conditioned on fitness scores. Another line attempts to propose mutations iteratively by greedy sampling \cite{sinai2020adalead}, reinforcement learning \cite{angermueller2019model}, or proximal gradients \cite{kirjner2023optimizing} on the learned fitness landscape. Differing from prior studies, MutaPLM incorporates textual instructions instead of fitness scores as navigation and proposes mutations satisfying human preferences.

%% file: 3_method.tex
\section{Methods}
\begin{figure*}[tpb]
\centering
    \includegraphics[width=0.94\linewidth]{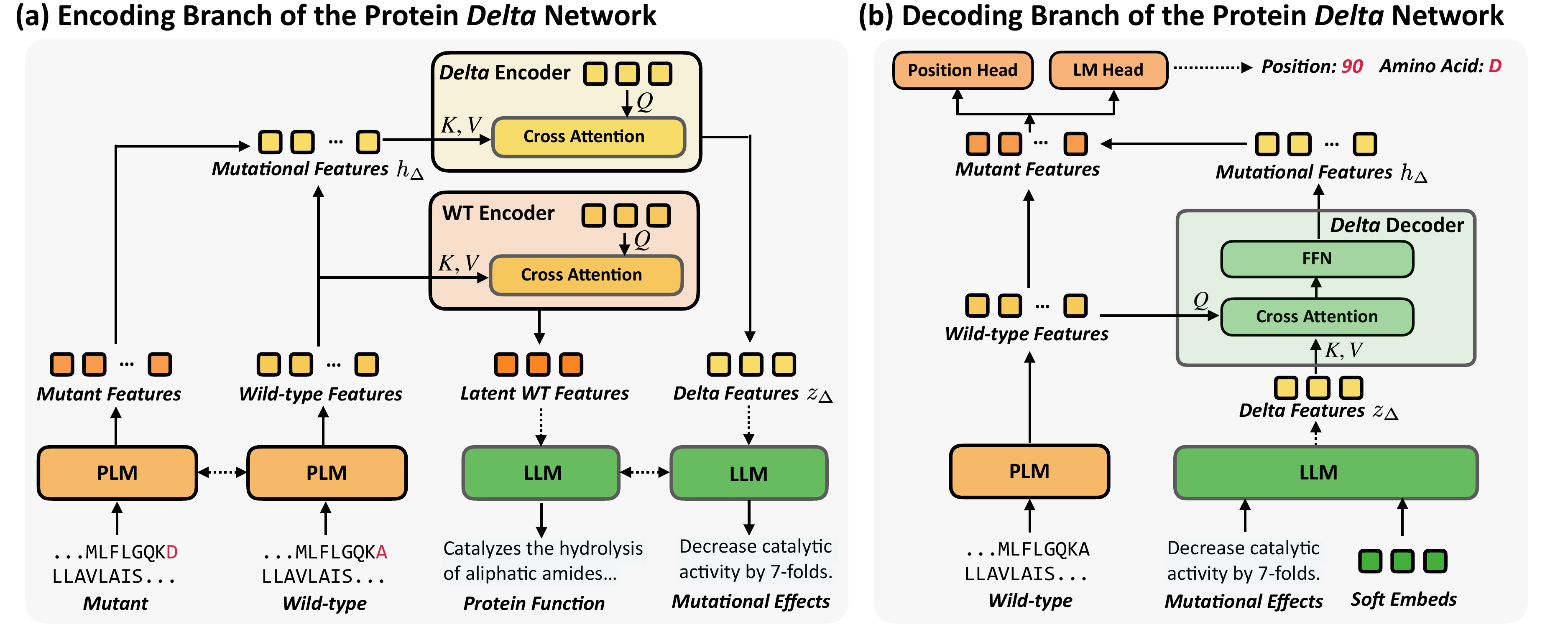}
\captionsetup{font={small,stretch=0.95}}
\caption{\textbf{Model architecture of MutaPLM. (a) The encoding branch of the protein \textit{delta} network.} The \textit{delta} encoder takes the subtraction of the PLM representations of the mutant and wild-type as inputs to generate $z_{\Delta}$. \textbf{(b) The decoding branch of the protein \textit{delta} network.} The key components involve a \textit{delta} decoder that reconstructs mutant features and two prediction heads deciding the position and amino acid of the mutation.}
\label{fig:model}
\vspace{-0.2cm}
\end{figure*}

The main goal of our work is to develop explainable and engineerable PLMs by explicitly modeling protein mutations. To achieve this goal, we elaborate on the proposed MutaPLM framework, highlighting three design components: (1) a protein \textit{delta} network that translates between mutations and protein \textit{delta} features $z_\Delta$ (Sec. \ref{sec:model}, detailed in Appendix \ref{app:model}), (2) a transfer learning pipeline with a chain-of-thought strategy that harvests protein mutation knowledge from cross-modal supervision (Sec. \ref{sec:pt}, detailed in Appendix \ref{app:train}), and (3) a specifically constructed dataset with diverse proteins and rich textual annotations of mutation effects (Sec. \ref{sec:dataset}, detailed in Appendix \ref{app:MutDes-description}). 
\subsection{Protein \textit{Delta} Network for Explicit Mutation Modeling \label{sec:model}}
The protein \textit{delta} network follows an encoder-decoder architecture, utilizing textual semantics as the latent feature space for protein mutations. As illustrated in Fig. \ref{fig:model}, the protein \textit{delta} network is composed of a protein language model (PLM), a large language model (LLM), a wild-type encoder, a \textit{delta} encoder, a \textit{delta} decoder, and two mutation prediction heads. We leverage ESM-2 (650M) \cite{lin2023evolutionary}, a powerful PLM pre-trained on evolutionary-scale databases, to encode protein sequences. We initialize the LLM with BioMedGPT-LM \cite{luo2023biomedgpt}, a scientific language model built on LLaMA2-7B \cite{touvron2023llama} through continual pre-training \cite{ke2022continual} on large-scale biomedical corpora. 

\textbf{Formulation of protein \textit{delta} features.} We speculate that the subtraction of PLM representations between the mutant and wild-type, denoted as $h_\Delta$, contains rich mutation information, making it suitable for extracting protein \textit{delta} features $z_\Delta$. Specifically:
\begin{equation}
    h_\Delta=h_{\text{mt}} - h_{\text{wt}}=f_{\text{PLM}}(x_{\text{mt}})-f_{\text{PLM}}(x_{\text{wt}}),
\end{equation}
where $x_{\text{mt}}$ and $x_{\text{wt}}$ are the amino acid sequences of the mutant and wild-type protein, $h_{\text{mt}}$ and $h_{\text{wt}}$ are their sequence representations, and $f_{\text{PLM}}$ is the protein language model. 

The \textit{delta} encoder $f_{\text{enc}}$ and \textit{delta} decoder $f_{\text{dec}}$ facilitates bi-directional transformations between $h_\Delta$ and $z_\Delta$ as follows:
\begin{equation}
    z_\Delta=f_{\text{enc}}(h_{\Delta}),\quad h_\Delta=f_{\text{dec}}(z_\Delta).
\end{equation}

\begin{figure*}[tpb]
\centering
    \includegraphics[width=0.99\linewidth]{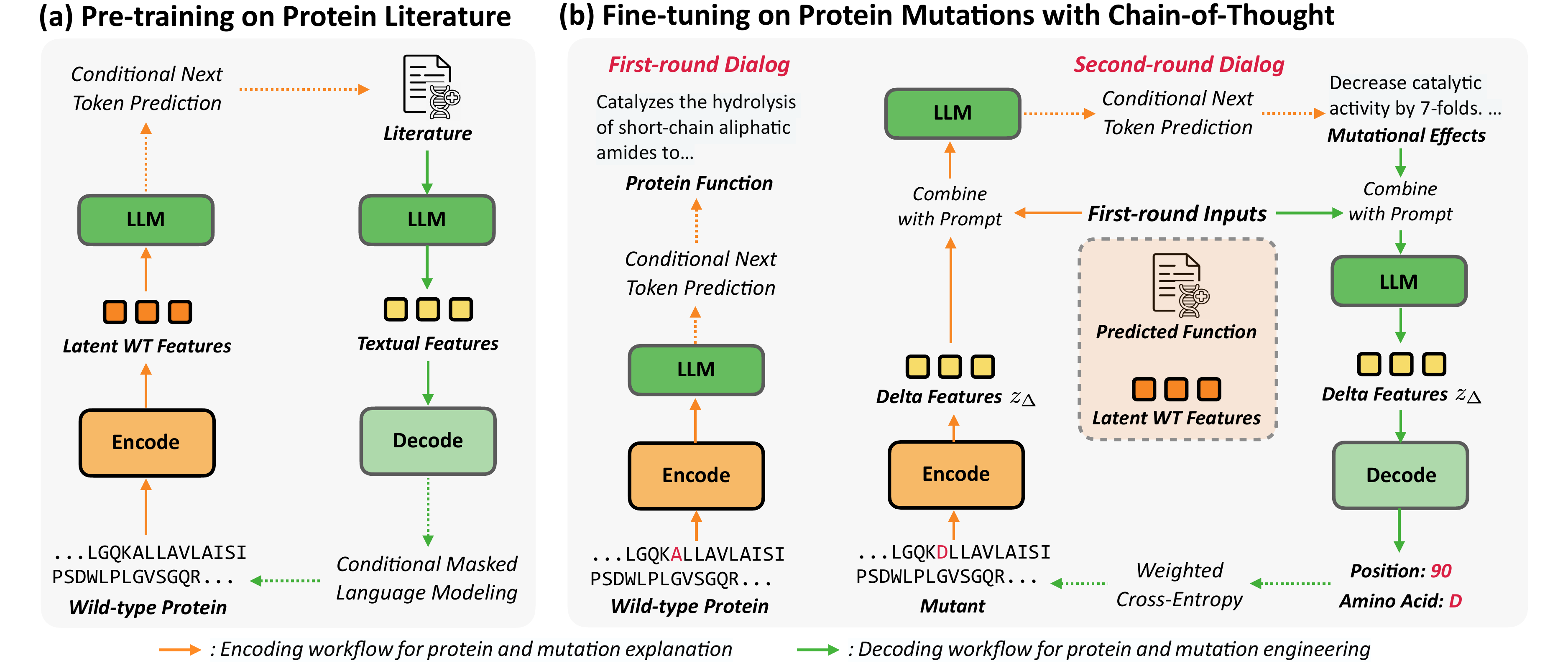}
\captionsetup{font={small,stretch=0.95}}
\caption{\textbf{Training pipeline of MutaPLM. (a) Workflow of pre-training on protein-related literature.} We perform next token prediction for the encoding workflow and conditional masked language modeling for the decoding workflow. \textbf{(b) Workflow of fine-tuning with chain-of-thought (CoT).} We employ a two-round dialog that involves describing the functions of a wild-type protein, explaining the effects of its mutation, and predicting the mutation based on the mutational effects.}
\label{fig:train}
\vspace{-0.2cm}
\end{figure*}

\textbf{Encoding protein \textit{delta} features.} Given $h_{\Delta}$, the \textit{delta} encoder is expected to extract information-preserving protein \textit{delta} features $z_\Delta$ within a unified feature space. However, protein sequences vary in length, ranging from several tens to thousands of amino acids. To address this issue, we adopt a cross-attention module \cite{vaswani2017attention} to transform the sequential representations into a fixed number of latent features. The module, partly inspired by BLIP series \cite{li2022blip,li2023blip}, maintains $K$ trainable features that serve as queries and takes the sequence representations as keys and values to generate outputs. We employ two parallel modules for encoding the wild-type features $h_\text{wt}$ and mutational features $h_{\Delta}$. 

\textbf{Decoding protein \textit{delta} features.} Drawing inspirations from LM-DESIGN \cite{zheng2023structure}, we introduce a cross-attention module that takes a symmetrical form of the \textit{delta} encoder. Specifically, it treats the wild-type protein representations $h_{\text{wt}}$ as queries and protein \textit{delta} features $z_{\Delta}$ as keys and values. The outputs are then processed by a two-layer feed-forward network (FFN) to reconstruct $h_{\Delta}$. The mutant representations $h_{\text{mt}}$ are obtained by combining $h_\Delta$ with $h_{\text{wt}}$, and fed into a position head and a language modeling head to predict the mutation. The position head is a fully-connected layer that predicts whether the amino acid should be substituted. The language modeling head is initialized from the PLM and predicts the type of the mutated amino acid. To facilitate text-based protein engineering, we maintain $K$ trainable soft tokens, which are appended to the input token embeddings of the LLM to summarize textual semantics. The output representations of the soft tokens are processed by the \textit{delta} decoder to generate mutations. 

Compared with previous works that connect protein sequences with LLMs \cite{fang2023mol, zhuo2024protllm, lv2024prollama}, the proposed protein \textit{delta} network exhibits the following advantages:
\begin{itemize}[noitemsep,topsep=0pt,parsep=4pt,partopsep=0pt,leftmargin=20pt]
    \item Explicit modeling of protein mutations. Prior models are designed for static protein sequences, while MutaPLM models the alterations introduced by mutations with protein \textit{delta} features $z_\Delta$.
    \item Encoder-decoder architecture. Prior works adopt either an encoder or a decoder architecture for protein sequences, while MutaPLM incorporates both encoding and decoding components.
\end{itemize}

\subsection{Transfer Learning with Cross-modal Supervision\label{sec:pt}}

Biomedical texts contain rich expert-annotated information on protein properties and mutational effects. As depicted in Fig. \ref{fig:train}, MutaPLM harvests these cross-modal supervision signals through a transfer learning pipeline, which we detail as follows: 

\textbf{Pre-training on protein literature.} In this stage, we aim to incorporate general protein knowledge from scientific publications with language modeling objectives, as shown in Fig. \ref{fig:train}(a). (1) For the encoding workflow, we take the output representations of the wild-type encoder as LLM inputs and calculate the next-token prediction objective \cite{radford2018improving} for generating descriptive texts. (2) For the decoding workflow, we employ the conditional masked language modeling (CMLM) objective \cite{ghazvininejad2019mask} on the protein sequence. Specifically, we mask 15\% amino acids and task the PLM to recover the masks based on the remaining amino acid sequence and the LLM-summarized textual representations. It is worth noting that in this stage, the \textit{delta} decoder acts as a modality translator, generating bias terms that help reconstruct the original sequence instead of capturing protein mutation information. Overall, we optimize the summation of these two language modeling objectives.

\textbf{Fine-tuning on protein mutations with chain-of-thought (CoT).} As depicted in Fig. \ref{fig:train}(b), we fine-tune MutaPLM on textual annotations of mutational effects to facilitate mutation explanation and engineering. Since mutational effects typically involve the enhancement or attenuation of protein functions, we adopt a chain-of-thought (CoT) strategy \cite{wei2022chain} that seamlessly connects protein functions and mutational effects within a two-round dialogue. In the first round, we prompt the LLM to describe the functions of the wild-type protein using the encoding workflow. In the second round, we introduce two tasks, namely describing the mutational effects with the encoding workflow, and predicting the mutation based on textual instructions with the decoding workflow. Both tasks utilize the latent wild-type representations and the predicted functions from the first round dialogue as additional inputs. Formally, the overall objective of fine-tuning is the summation of three parts: (1) next token prediction on protein function descriptions, (2) next token prediction on mutational effects, and (3) weighted cross-entropy between the predicted mutation and the ground-truth mutation. 




\begin{table}[htpb]
\captionsetup{font={small,stretch=0.95}}
\captionof{table}{\textbf{Statistics of the MutaDescribe dataset.} We report the number of proteins and samples, the average protein sequence length, and the average number of words for mutational effects.}
\label{tab:dataset}
\centering
\begin{tabular}{lcccc}
\toprule
Split   & \# Proteins & \# Samples & Avg. sequence length & Avg. words \\
\midrule
Train         & 20,553      & 165,236    & 516.1       & 28.3       \\
Valid         & 2,207       & 4,663      & 524.8       & 28.3       \\
Test (Easy)   & 429         & 460        & 518.1       & 27.3       \\
Test (Medium) & 68          & 384        & 669.6       & 31.6       \\
Test (Hard)   & 81          & 404        & 530.0       & 31.8       \\
\bottomrule
\end{tabular}
\end{table}

\subsection{MutaDescribe: A Diverse Protein Mutation Dataset with Textual Annotations \label{sec:dataset}}
We build MutaDescribe, a large-scale dataset comprising 20.9K wild-type proteins and 171.1K single-site mutations, to facilitate fine-tuning and evaluation. We provide an overview of our dataset in Tab. \ref{tab:dataset}. The construction process involves the following steps:

\textbf{Raw data collection.} The primary source of MutaDescribe is UniProtKB/SwissProt \cite{boutet2016uniprotkb}, a widely adopted protein database that contains 106.6K single-site substitutions. We collect expert-reviewed descriptions of mutational effects from the \textit{Phenotypes \& Variants} entry and retrieve the abstract of the corresponding publications on PubMed \cite{canese2013pubmed} based on available reference information.

\textbf{Quality control.} We prompt GPT-3.5-turbo \cite{achiam2023gpt} to filter out low-quality descriptions such as those that only mention the originating species. This step helps ensure that the dataset contains high-quality and informative annotations.

\textbf{Data enrichment.} Given that the descriptions in UniProtKB are generally short and homogeneous, we utilize GPT-3.5-turbo to enrich the textual annotations by retrieving relevant descriptions from the original PubMed abstract. Additionally, we balance the number of benign and malignant mutations by constructing reversed samples. Specifically, for each mutation, we attempt to exchange the wild-type and the mutant and prompt GPT-3.5-turbo to write a description opposite to the original mutational effect. For example, if the mutational effect of an A89H mutation is \textit{"Increased catalytic activity"}, we will create a reversed sample with an H89A mutation and \textit{"Decreased catalytic activity"}.

\textbf{Data splitting.} We first randomly split our dataset into training, validation, and test sets. To evaluate models' generalization capabilities on novel proteins, we further partition the test set into three subsets based on the wild-type sequence homology with training sequences. We adopt MMSeqs2 \cite{steinegger2017mmseqs2}, a widely-adopted tool to calculate sequence homology. The \textit{Easy, Medium} and \textit{Hard} test subsets comprise samples whose sequence homology are between $[0.95,1], [0.5, 0.95)$, and $[0, 0.5)$ respectively. We also implement a temporal split based on the publication date of the mutation, and we defer readers to Appendix \ref{app:data} for details and Appendix \ref{app:temporal-split} for evaluation results.

Compared with prior mutation benchmarks \cite{landrum2018clinvar, riesselman2018deep, notin2024proteingym}, MutaDescribe is the first to incorporate textual annotations for facilitating mutation explanation and engineering. Besides, MutaDescribe contains a wider variety of wild-type proteins, surpassing ProteinGym \cite{notin2024proteingym} by 6 times in quantity.

%% file: 4_experiments.tex
\section{Experiments\label{sec:exp}}

\begin{table}[tpb]
\centering
\captionsetup{font={small,stretch=0.95}}
\caption{\textbf{Performance evaluation for mutation explanation on the test sets of MutaDescribe.} R-L: ROUGE-L. BL-2: BLEU-2.}
\setlength\tabcolsep{4pt}
\begin{tabular}{lcccccccc}
\toprule
\multirow{2}{*}{Model} & \multicolumn{2}{c}{Easy} & \multicolumn{2}{c}{Medium} & \multicolumn{2}{c}{Hard} & \multicolumn{2}{c}{Average} \\
 & R-L &  BL-2   & R-L  & BL-2  & R-L & BL-2 & R-L & BL-2  \\
\midrule
ProLLaMA \cite{lv2024prollama}       & 1.02 & 0.64 & 1.00 &  0.91  & 1.03 & 0.70 & 1.02 & 0.74  \\
Mol-Instructions \cite{fang2023mol} & 5.10 & 0.65 & 5.19 &  0.65  & 5.56 & 0.90 & 5.27 & 0.73 \\
Galactica-6.7B \cite{taylor2022galactica} & 6.53  & 3.52 & 7.64  &  3.58  & 7.33 & 2.88 & 7.13 & 3.33 \\
\midrule
GPT-4-0613 (1-shot) \cite{achiam2023gpt}     & 8.04   &  2.93  &  9.96  &  3.42  &  9.62  &  2.69 & 9.14 & 3.00 \\
GPT-4-0613 (5-shot) \cite{achiam2023gpt}      & 10.46  &  2.51  &  10.31 &  2.81  & 10.79  &  1.88 & 10.52 & 2.40 \\
GPT-4-0613 (5-shot, kNN) \cite{achiam2023gpt} & 11.63  &  9.63  &  12.98 &  10.88 & 12.46  &  8.63 & 12.31 & 9.69  \\
GPT-4 + ESM-2 \cite{lin2023evolutionary} & 11.69 & 11.09 & 13.02 & 11.50 & 12.77 & 8.48 & 12.45 & 10.37 \\
GPT-4 + OntoProtein \cite{zhang2021ontoprotein} & 11.84 & 10.93 & 12.69 & 11.22 & 12.81 & 8.17 & 12.42 & 10.13 \\
\midrule
AugmentedESM \cite{hsu2022learning}     & 11.60 & 8.33 & 11.40 & 7.46 & 10.73 & 6.95 & 11.26 & 7.62 \\
Fine-tuned ESM-2 \cite{lin2023evolutionary}  & 20.49 & 9.37 & 11.87 & 5.95 & 11.34 & 3.32 & 14.88 & 6.36 \\
\midrule
MutaPLM        & \textbf{25.80} & \textbf{18.77} & \textbf{21.07} & \textbf{12.59}  &  \textbf{16.51} & \textbf{8.69} & \textbf{21.34} & \textbf{13.61} \\
\bottomrule
\end{tabular}
\label{tab: explanation_result}
\end{table}

\begin{figure}[tpb]
    \centering
    \includegraphics[width=\linewidth]{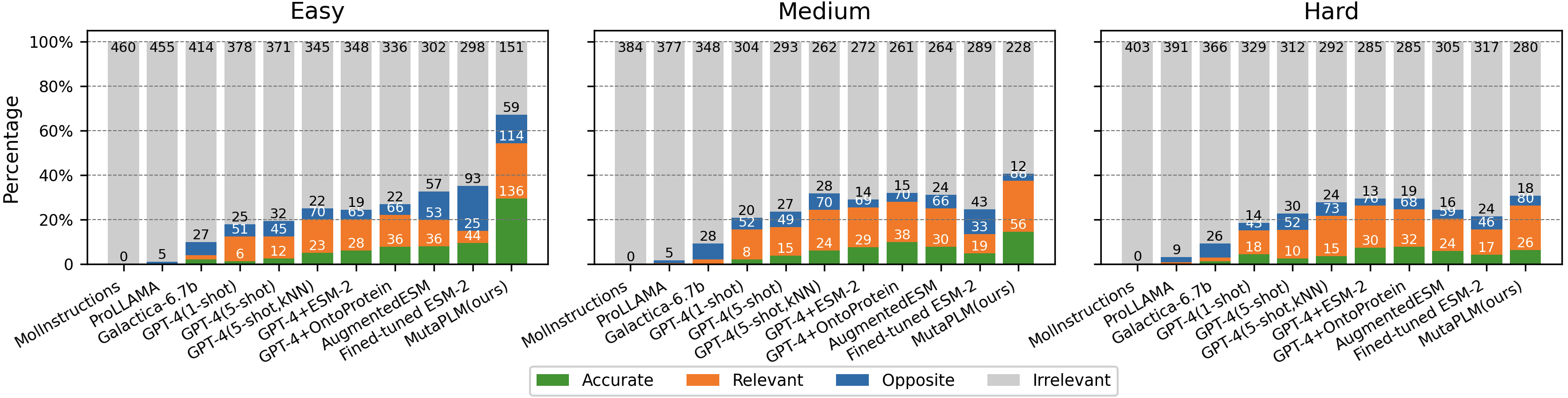}
    \captionsetup{font={small,stretch=0.95}}
    \caption{\textbf{Human-AI collaborative evaluation results for mutation explanation on the test sets of MutaDescribe.} We show the number of accurate, relevant, opposite, and irrelevant predictions.}
    \label{fig:gpt-eva}
\end{figure}

In this section, we demonstrate that MutaPLM is adept at interpreting and engineering mutations through comprehensive experiments. We start with a brief introduction of our training setups (Sec. \ref{exp:setup}), followed by detailed evaluations on two core tasks: mutation explanation (Sec. \ref{exp:exp}) and mutation engineering (Sec. \ref{exp:eng}). We also present an in-depth analysis of our design components (Sec. \ref{exp:abla}), including pre-training and the CoT strategy.
\subsection{Training Setup\label{exp:setup}}
To alleviate catastrophic forgetting \cite{luo2023empirical} and save computational costs, we train MutaPLM in a parameter-efficient way. We apply low-rank adaptation (LoRA) \cite{hu2021lora} on the LLM with a rank of 16. The number of query embeds and soft tokens is set as $K=32$. We optimize the LoRA modules, the wild-type encoder, the \textit{delta} encoder, the \textit{delta} decoder, the soft tokens, the position head, and the language modeling (LM) head, which comprises a total of 75.0M parameters. The remaining 7.4B parameters are kept frozen.

We pre-train MutaPLM for 200K steps with a batch size of 32 on 1.1M protein-text data collected from biomedical publications (detailed in Appendix \ref{app:pt_data}) and fine-tune it for 70K steps with a batch size of 24 on MutaDescribe. For both stages, we use the AdamW optimizer \cite{loshchilov2018decoupled} with a learning rate that is linearly warmed up to $10^{-4}$ for the first 1K steps and decreases to $10^{-5}$ following a cosine annealing strategy. The overall training process takes 10 days on 4 NVIDIA A100 GPUs.

\subsection{Performance Evaluation on Mutation Explanation\label{exp:exp}}

\begin{figure}[tpb]
    \centering
    \includegraphics[width=\textwidth]{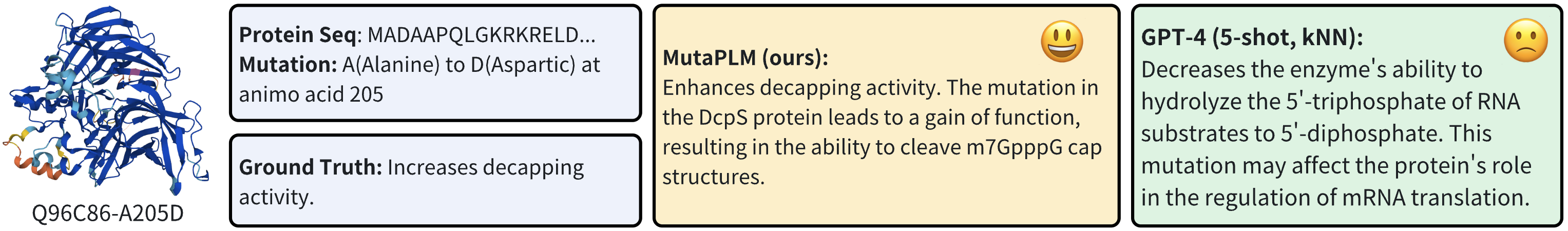}
    \captionsetup{font={small,stretch=0.95}}
    \caption{\textbf{Case study for a mutation from A (Alanine) to D (Aspartic) at the 205-th position of \textit{m7GpppX diphosphatase}.} MutaPLM provides accurate explanations and insights, while GPT-4 generates irrelevant results.}
    \label{fig:case-study}
    \vspace{-0.2cm}
\end{figure}

Differing from existing studies that interpret mutational effects with protein fitness \cite{meier2021language, zhao2024contrastive}, we formulate mutation explanation as providing detailed textual descriptions for protein mutations.

\textbf{Baselines.} While no prior work is specifically designed for this task, we perform zero-shot analysis on popular LLMs with various zero-shot or few-shot prompts and implement supervised models for comparison. Our baselines include (1) Text-based LLMs. We perform in-context learning \cite{dong2022survey} by providing 1-shot and 5-shot demonstrations to GPT-4 \cite{achiam2023gpt}, the most advanced model in NLP. Additionally, we implement a k-nearest neighbor (kNN) strategy \cite{ai4science2023impact} that selects the top-k homologous proteins from the training set as few-shot examples. (2) LLM-assisted PLMs, including ESM-2 \cite{lin2023evolutionary} and OntoProtein \cite{zhang2021ontoprotein}. In addition to kNN-based 5-shot samples for GPT-4, we leverage PLMs to provide additional information by predicting the evolutionary plausibility of the mutation. (3) LLMs trained on protein sequences, including Galactica-6.7B \cite{taylor2022galactica}, Mol-Instructions \cite{fang2023mol}, and ProtLLM \cite{zhuo2024protllm}. We feed the wild-type and mutated protein sequences into these models and instruct them to provide mutation explanations. (4) Fine-tuned LLMs. We fine-tune BioMedGPT-LM by feeding the ESM-2 representations of the wild-type and mutant (Fine-tuned ESM-2) or the wild-type sequence and evolutionary plausibility (AugmentedESM \cite{hsu2022learning}) into the LLM and performing casual generation. Notably, for all ESM-2 models used in our baselines, we adopt the model with 650M parameters for fair comparison. We defer readers to Appendix \ref{app: baseline_setup} for more implementation details.


\textbf{Evaluation.} We adopt BLEU \cite{papineni2002bleu} and ROUGE \cite{lin2004rouge} scores to assess the quality of the generations by comparing them with ground-truth annotations. To further investigate whether the predictions are truly insightful and helpful in studying protein mutations, we perform a human-AI collaborative evaluation. Specifically, we first utilize GPT-4 as a proxy of human experts to categorize the predictions into \textit{Accurate}, \textit{Relevant}, \textit{Opposite}, and \textit{Irrelevant}, based on the relevance between the predictions and ground truth. Then, we recruit a postgraduate from a top university who majors in biology to assess and rectify GPT-4 evaluation results on mutation explanations following the same categorization protocol. The prompt and detailed evaluation results are displayed in Appendix \ref{app: prompt_eva}.

\textbf{Results and analysis.} We present performance comparisons on the test sets of MutaDescribe in Tab. \ref{tab: explanation_result} and Fig. \ref{fig:gpt-eva}. We observe that: (1) MutaPLM achieves state-of-the-art performance across various evaluation metrics, outperforming fine-tuned ESM-2 by 6.46\% in ROUGE-L and GPT-4 + ESM-2 by 3.24\% in BLEU-2. Additionally, more than 40.22\% of MutaPLM predictions are regarded as \textit{Accurate} or \textit{Relevant} with ground-truth labels, which showcases our model's helpfulness in real-world research scenarios. (2) While the performance on the \textit{Medium} and \textit{Hard} sets is not as promising as in the easy set, MutaPLM shows generalization capabilities on novel proteins, as validated by 6.44\% accurate and 19.80\% relevant predictions on the hard set. (3) The evolutionary plausibility values are beneficial for elucidating mutational effects, as demonstrated by the slightly improved results of LLM-assisted PLMs against the plain GPT-4 counterpart. However, the superior performance of fine-tuned ESM-2 and MutaPLM indicates that integrating the mutant sequence provides richer mutational information. (4) Supervised baselines underperform few-shot GPT-4 models, especially on \textit{Medium} and \textit{Hard} sets and BLEU-2 scores. We observe that supervised models tend to randomly combine short textual segments from the training set, indicating overfitting problems. (5) LLMs trained on protein sequences perform poorly, as they are solely instruction-tuned on single protein sequences. Hence, we emphasize the significance of knowledge transfer from protein functions to mutational effects and their basic properties.

\textbf{Case study.} Additionally, we present a case study in Fig. \ref{fig:case-study} for a mutation from \textit{m7GpppX diphosphatase}. Our model accurately identifies the increased decapping activity and provides novel insights beyond the ground truth. In contrast, the GPT-4 model mistakenly identifies the mutational effects as decreases in enzymic activity. More cases are available in Appendix \ref{app:detailed-result}.




\subsection{Performance Evaluation on Mutation Engineering\label{exp:eng}}
\begin{table}[tpb]
\setlength\tabcolsep{2.5pt}
\captionsetup{font={small,stretch=0.95}}
\centering
\caption{\textbf{Performance evaluation for mutation engineering on the test sets of MutaDescribe.} Acc: prediction accuracy of the amino acid given the position of the mutation. Rec@50: top 50 recall of the desired mutant. -: not reported due to unaffordable computation costs (requires $\sim$ 1M forward passes).}
\label{tab:mut_design}
\begin{tabular}{lcccccccc}
\toprule
\multirow{2}{*}{Model} & \multicolumn{2}{c}{Easy} & \multicolumn{2}{c}{Medium} & \multicolumn{2}{c}{Hard} & \multicolumn{2}{c}{Average} \\
                       & Acc      & Rec@50     & Acc       & Rec@50      & Acc      & Rec@50     & Acc       & Rec@50      \\
\midrule
Random    &  5.23   & 0.83  &   4.94   &   0.52   &  5.20   &  1.24  &   5.13 &   0.87 \\
\midrule
ProtST (ESM-2) \cite{xu2023protst}         &  5.86  &  -  & 6.51 & - & 7.18  & - & 6.49 & -   \\
\midrule
GPT-4-0613 (1-shot) \cite{achiam2023gpt}   &  10.83  &  5.00  &  10.77  &  6.92  & 12.09  & 8.79   &  11.21  & 6.81 \\
GPT-4-0613 (5-shot) \cite{achiam2023gpt}   &  14.84  &  4.68  &  9.32   &  6.78  & 13.33  & 5.62  & 12.65 & 5.63 \\
GPT-4-0613 (5-shot, kNN) \cite{achiam2023gpt}   & 15.97 & 7.56 &  14.29  &  7.14  & 14.77  & 7.95 &  15.06 & 7.56 \\
\midrule
ESM-2 \cite{lin2023evolutionary}      &  35.21   & 23.91  & 34.63  &  22.91   &  37.87  & 28.71 & 35.84 & 25.15    \\
OntoProtein  \cite{zhang2021ontoprotein}   &  39.78  &  28.91  &  36.45  & 26.04  & 38.61  &  29.20 & 38.37 & 28.12   \\
\midrule
Fine-tuned BioMedGPT \cite{luo2023biomedgpt}     & 35.21   &  7.82  & 32.29   & 5.72  & 39.60  & 12.62  & 35.73 & 8.72   \\
Fine-tuned ESM-2 \cite{lin2023evolutionary, gu2021domain}     &  52.17   & 35.65  & \textbf{52.08}  &  30.60   &  50.00  & 34.65 & 51.43 & 33.77    \\
\midrule
MutaPLM          & \textbf{56.08}  & \textbf{43.47}   &  48.69  & \textbf{34.89}  & \textbf{55.19}  &  \textbf{43.81} & \textbf{53.51} & \textbf{40.94} \\
\bottomrule
\end{tabular}
\end{table}

Differing from prior works \cite{sinai2020adalead, angermueller2019model, kirjner2023optimizing} that perform mutation engineering with an active learning paradigm \cite{ren2021survey}, we challenge models to directly propose protein mutations based on the wild-type sequence and textual instructions. As we primarily focus on single-site mutations, we formulate this as a retrieval task from $19\times L$ possible mutants for a protein sequence of length $L$.

\textbf{Baselines.} We adopt four groups of baselines including: (1) Few-shot LLMs. Similar to mutation explanation, we prompt GPT-4 to suggest single-site mutations through in-context few-shot learning. (2) Zero-shot PLMs including ESM-2 \cite{lin2023evolutionary} and OntoProtein \cite{zhang2021ontoprotein}. We calculate the evolutionary plausibility scores following \cite{meier2021language} for each amino acid and derive the best mutant. (3) A retrieval-based model, namely ProtST (ESM-2) \cite{xu2023protst}. We calculate the cosine similarity between PLM and textual representations of mutational effects to score and rank mutations. (4) Fine-tuned models. We fine-tune BioMedGPT \cite{luo2023biomedgpt} to directly propose a mutation based on the protein sequence and instruction. We also fine-tune ESM-2 by combining its wild-type sequence representations with  BioMedBERT \cite{gu2021domain} encodings of textual instructions by a cross-attention layer. Please refer to Appendix \ref{app: baseline_setup} for details of our baselines.

\textbf{Evaluation.} We report the average accuracy of the mutated amino acid on the ground-truth mutational position. We also report top-50 recall scores on all possible mutations.

\textbf{Results and analysis.} Comparisons between MutaPLM and baselines on the test sets of MutaDescribe are presented in Tab. \ref{tab:mut_design}. We observe that: (1) MutaPLM achieves an average of 53.51\% in accuracy and 40.94\% in top-50 recall, improving the original ESM-2 model by 1.6-fold. The substantial gains of MutaPLM underscore the significance of textual navigation in mutation engineering. (2) MutaPLM outperforms the fine-tuned ESM-2 model by an average of 2.09\% in accuracy and 6.17\% in top-50 recall, benefiting from our architectural design and pre-training. (3) The overall performance of MutaPLM on the \textit{Easy} and \textit{Hard} sets are similar but significantly higher than on the \textit{Medium} set. We attribute this to data distribution: protein sequences in the \textit{Medium} set are longer (see Tab. \ref{tab:dataset}), and the distribution of the mutated amino acids differs (see Fig. \ref{fig:app-dataset}). Besides, the PLM may have witnessed the wild-type protein during pre-training, which mitigates the overfitting problem. (4) Compared to LLMs, both zero-shot and fine-tuned PLMs achieve superior performance, thanks to their evolutionary knowledge attained from pre-training on large-scale protein sequences. (5) Aligning the representations of protein sequences and texts cannot benefit mutation modeling, as evidenced by the poor performance of ProtST (ESM-2).

\begin{figure*}[tpb]

\centering
    \includegraphics[width=0.83\linewidth]{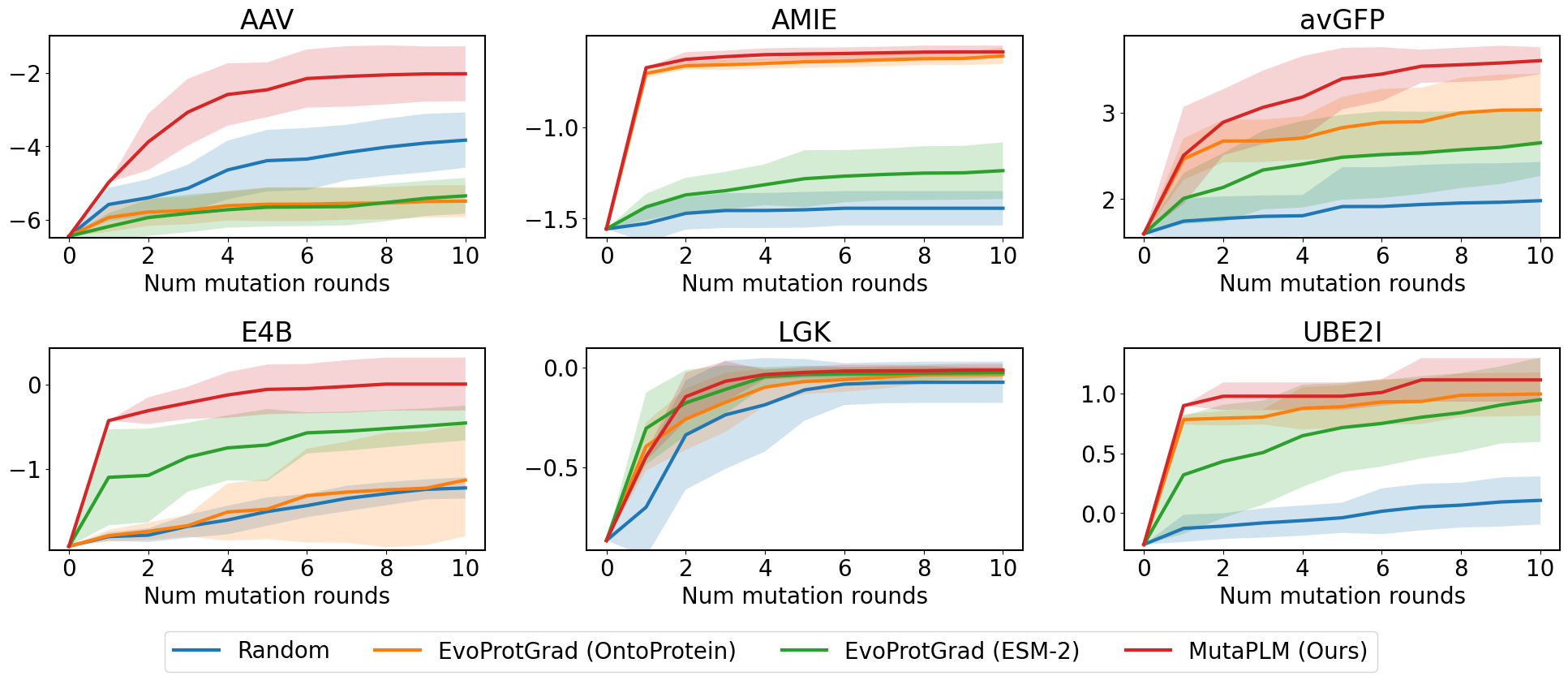}
\captionsetup{font={small,stretch=0.95}}
\caption{\textbf{Visualization of fitness scores for multi-round protein optimization.} The curves indicate the average results, and the shaded regions indicate the standard deviation.}
\vspace{-0.3cm}
\label{fig:fitness}
\end{figure*}

\textbf{Visualization of protein fitness on multi-round optimization.} In addition to single-site mutations, we employ a beam-search algorithm \cite{freitag2017beam} to obtain multi-point substitutions iteratively. We manually write the optimization objective for 6 representative benchmarks, set the number of beams as 20, perform 20 independent runs, and visualize the fitness scores predicted by ESM landscape models \cite{ren2022proximal}. We compare MutaPLM with EvoProtGrad \cite{emami2023plug}, a gradient-based strategy that leverages PLMs for multi-round optimization, as well as with random sampling. More details are presented in Appendix \ref{app:optimize}. As shown in Fig. \ref{fig:fitness}, our model consistently yields higher-fitness mutants across 6 proteins with varying objectives, especially in the initial rounds of optimization. These results highlight MutaPLM's potential in assisting real-world mutagenesis applications. 

\subsection{In-depth Analysis\label{exp:abla}}
\textbf{Impacts of transfer learning.} We show the impacts of pre-training and fine-tuning in Fig. \ref{fig:steps}. As the fine-tuning proceeds, the performance of MutaPLM continues to improve on the \textit{Easy} set but deteriorates on the \textit{Medium} and \textit{Hard} sets, indicating overfitting problems on out-of-domain samples. Besides, without pre-training, MutaPLM achieves higher performance for the initial steps, which we attribute to the adaptation cost from pre-training texts to fine-tuning texts. However, the overall ROUGE-L scores decline by 1.56\% for mutation explanation and 1.18\% for mutation engineering as the fine-tuning finalizes. Overall, these results validate our transfer learning design. 

\textbf{Impacts of chain-of-thought (CoT).} To investigate the impacts of the chain-of-thought strategy, we perform ablation studies by (1) replacing the predicted function with the ground truth description, (2) replacing the predicted function with \textit{'Unknown function'}, (3) removing the \textit{delta} features for mutation explanation, and (4) removing the mutational effects for mutation engineering. As shown in Tab. \ref{tab:ablation}, removing protein functions leads to a performance drop of 2.80\% for mutation explanation and 1.13\% for mutation engineering. Conversely, using the ground-truth function results in notable improvements, particularly for mutation explanation. Besides, the \textit{delta} features and mutational effects within the second-round dialog play more significant roles in MutaPLM. These findings highlight the significance of jointly incorporating protein function and mutation information in explaining and navigating protein mutations.

\begin{minipage}[htpb]{0.4\textwidth}
\centering
\vspace{-1em}
\small
\captionsetup{font={small,stretch=0.95}}
\captionof{table}{\textbf{Ablation studies.} w/o: without. w/: with. We report average ROUGE-L for mutation explanation and average Recall@50 for mutation engineering.}
\label{tab:ablation}
\setlength\tabcolsep{2.5pt}
\begin{tabular}{lcc}
\toprule
 Model & Explain & Engineer \\
\midrule
 MutaPLM   & 21.34 & 40.94  \\
 \midrule
  w/ golden function   & 23.80 & 41.26    \\
 w/o function   & 18.54 & 39.81    \\
 w/o \textit{delta} features & 17.36 & - \\
 w/o mutational effects & -  & 35.10 \\
\bottomrule
\end{tabular}
\end{minipage}
\hspace{0.3em}
\begin{minipage}[htpb]{0.57\textwidth}
\centering
    \includegraphics[width=\linewidth]{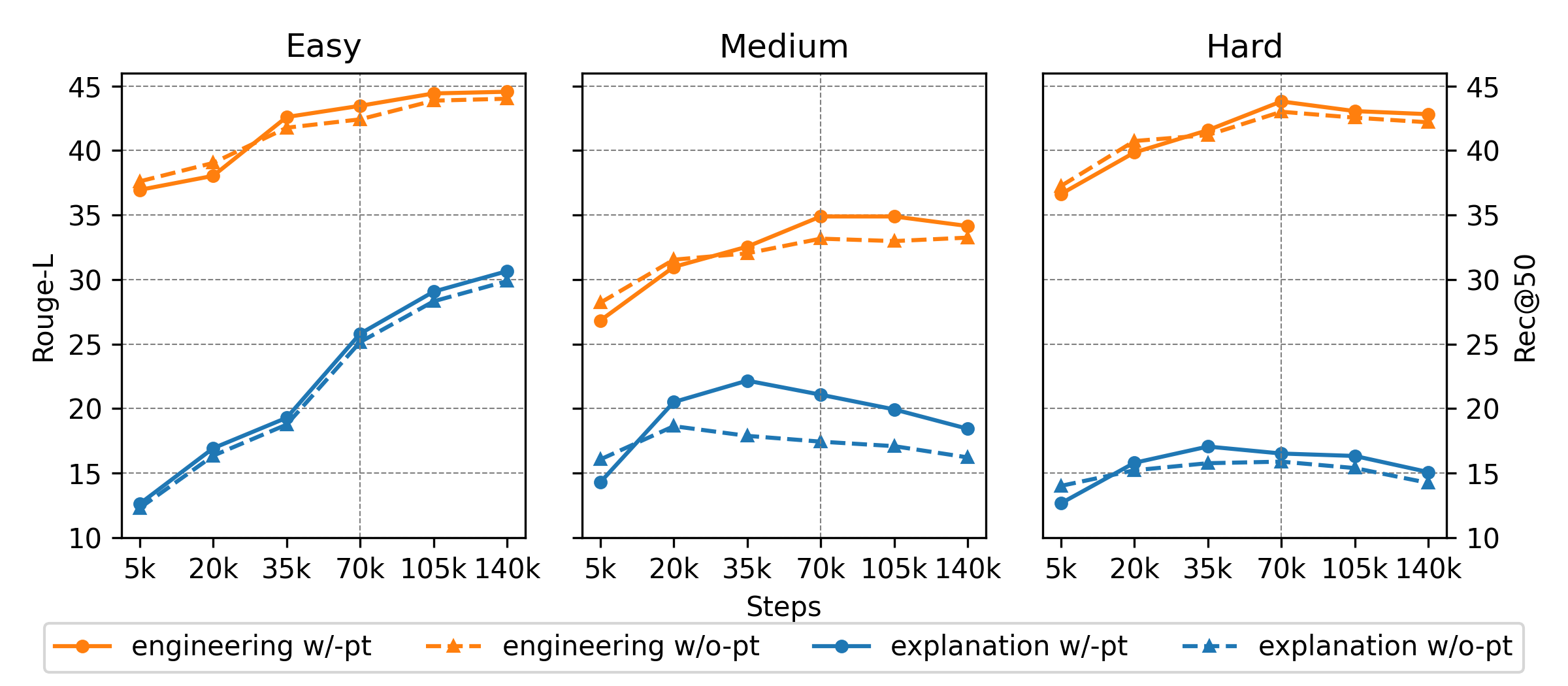}
\captionsetup{font={small,stretch=0.95}}
\captionof{figure}{\textbf{Performance analysis for mutation explanation (blue) and engineering (orange) on pre-training and fine-tuning.} w/o pt: without pre-training. w/ pt: with pre-training.}
\label{fig:steps}
\end{minipage}

%% file: 5_conclusion.tex
\section{Limitations and Broader Impacts \label{sec:lim}}
MutaPLM pioneers as the first attempt in the explicit modeling of protein mutations with natural language, and we expect future endeavors on (1) expanding the scale and diversity of the MutaDescribe dataset by integrating multi-point mutations and indels \cite{notin2024proteingym}, (2) analyzing the alterations of protein 3D structures \cite{abramson2024accurate} to deepen the understanding of mutations, and (3) developing active learning \cite{ren2021survey} pipelines to harness feedbacks from wet-lab experiments in real-world mutagenesis studies.

While MutaPLM bears promise in mutation explanation and engineering, we emphasize safety concerns that it can be misused to generate pathogenic mutations and harmful bio-agents. Hence, we declare that MutaPLM, upon public release, should be restricted to research purposes, and any further applications should undergo comprehensive experiments and human inspections. 

\section{Conclusions}
In this work, we present MutaPLM, a unified framework harvesting protein language models for mutation explanation and engineering. We propose a protein \textit{delta} network to model mutations explicitly with protein \textit{delta} features and develop a transfer learning pipeline with a chain-of-thought strategy to integrate protein mutation knowledge from biomedical texts. Additionally, we construct MutaDescribe, the first large-scale dataset containing diverse proteins and detailed textual annotations for mutations. Our experiments demonstrate that MutaPLM offers insightful explanations for mutational effects and proposes desirable mutants based on textual instructions. We anticipate that the proposed MutaPLM framework and our publicly released dataset will pave the way for novel research avenues and applications in studying proteins.



%% file: 6_appendix.tex
\setcounter{section}{0}
\setcounter{equation}{0}
\setcounter{subsection}{0}
\setcounter{table}{0}
\setcounter{figure}{0}
\renewcommand{\theequation}{A\arabic{equation}}
\renewcommand{\thefigure}{A\arabic{figure}}
\renewcommand{\thetable}{A\arabic{table}}
\renewcommand{\thesubsection}{\Alph{subsection}}
\newpage
\section*{Appendix}

\subsection{Details of MutaPLM}
\subsubsection{Model Architecture \label{app:model}}
Our protein \textit{delta} network consists of a protein language model (PLM), a large language model (LLM), a wild-type encoder, a \textit{delta} encoder, a \textit{delta} decoder, and two prediction heads for mutation. We introduce these components as follows:

\textbf{Protein language model.} We formulate the wild-type protein as an amino acid sequence $x_{\text{wt}}=\left[x^{(\text{wt})}_1, x^{(\text{wt})}_2, \cdots, x^{(\text{wt})}_L\right]$ of length $L$. We focus on single-site substitution mutants, denoted by its sequence $x_{\text{mt}}=\left[x^{(\text{mt})}_1, x^{(\text{mt})}_2, \cdots, x^{(\text{mt})}_L\right]$ satisfying $\mathcal{H}(x_{\text{wt}}, x_{\text{mt}})=1$, where $\mathcal{H}(\cdot,\cdot)$ is the Hamming distance. We adopt ESM-2 (650M) \cite{lin2023evolutionary} as our protein language model $f_{\text{PLM}}$, which transforms the protein sequences into dense feature vectors as follows:
\begin{equation}
\begin{aligned}
    h_{\text{wt}}=\left[h^{(\text{wt})}_1,h^{(\text{wt})}_2,\cdots,h^{(\text{wt})}_L\right]=f_{\text{PLM}}(x_{\text{wt}}),\\
    h_{\text{mt}}=\left[h^{(\text{mt})}_1,h^{(\text{mt})}_2,\cdots,h^{(\text{mt})}_L\right]=f_{\text{PLM}}(x_{\text{mt}}).\\
\end{aligned}
\end{equation}
 Then, we introduce the mutational representation, $h_\Delta$, calculated as follows:
\begin{equation}
    h_{\Delta}=\left[h_1^{(\Delta)},h_2^{(\Delta)},\cdots,h_L^{(\Delta)}\right]=h_{\text{mt}}-h_{\text{wt}}.
\end{equation}
\textbf{Large language model.} Similarly, we formulate biomedical texts as a sequence of tokens $t=[t_1,t_2,\cdots,t_N]$. We initialize our LLM with BioMedGPT-LM \cite{luo2023biomedgpt}, which is obtained by continually pre-training LLaMA2-7B \cite{touvron2023llama} on biomedical corpus. The large language model $f_{\text{LLM}}$ takes the following steps to transform $t$ into latent features and output distributions of the next token:
\begin{equation}
\label{equ:text}
\begin{aligned}
    &e=[e_1,e_2,\cdots,e_N]=g_{\text{emb}}(t),\\
    &z_t=[z_1,z_2,\cdots,z_N]=g_{\text{transformers}}(e),\\
    &P(t_i|t_{<i})=g_{\text{LM}}(h_i), \\
\end{aligned}
\end{equation}
where $g_{\text{emb}}$ is the word embedding layer, $z_t$ is the textual representation calculated by transformer blocks $g_{\text{transformers}}$, $g_{\text{LM}}$ is the language modeling head, and $P(t_i|t_{<i})$ is the probability distribution of $i$-th token based on preceding tokens.

\textbf{Wild-type encoder.} The wild-type encoder comprises $K$ trainable query vectors $q_{\text{wt}}=[q_1,q_2,\cdots, q_K]$ and a cross attention module. It transforms the wild-type representations $h_{\text{wt}}$ into a fixed number of features as follows:
\begin{equation}
\begin{aligned}
\label{equ:ca}
&z_{\text{wt}}=\left[z^{(\text{wt})}_1,z^{(\text{wt})}_2,\cdots,z^{(\text{wt})}_K\right]=\text{CrossAttention}_{\text{wt}}(q_{\text{wt}},h_{\text{wt}},h_{\text{wt}}),\\
&\text{CrossAttention}(Q,K,V)=\text{Softmax}\left(\frac{\hat{Q}\hat{K}^T}{\sqrt{d_k}}\right)\hat{V}\\
&\hat{Q}=QW^{Q}, \hat{K}=KW^{K}, \hat{V}=VW^{V},\\
\end{aligned}
\end{equation}
where $W^Q,W^K,W^V$ are trainable parameters, and $d_k$ is the feature dimension.

\textbf{\textit{Delta} encoder.} The \textit{delta} encoder follows the same architecture as the wild-type encoder. It encodes the protein \textit{delta} features as follows:
\begin{equation}
z_{\Delta}=\left[z^{(\Delta)}_1,z^{(\Delta)}_2,\cdots,z^{(\Delta)}_K\right]=\text{CrossAttention}_{\text{enc}}(q_{\Delta},h_{\Delta},h_{\Delta}),
\end{equation}
where $q_\Delta$ are the $K$ trainable queries, and the cross attention is calculated following Equ. \ref{equ:ca}. Notably, the wild-type encoder and \textit{delta} encoder comprise independent parameters.

\textbf{\textit{Delta} decoder.} The \textit{delta} decoder transforms the protein \textit{delta} features $z_\Delta$ back to the original mutation representations $h_\Delta$. It comprises a cross-attention layer and a two-layer feed-forward network with ReLU activation. Specifically:
\begin{equation}
\begin{aligned}
\label{equ:delta_dec}
    &\tilde{z}_\Delta=\text{CrossAttention}_{\text{dec}}(h_{\text{wt}},z_\Delta,z_\Delta),\\
    &h_\Delta=\text{FeedForward}(\tilde{z}_\Delta),
\end{aligned}
\end{equation}
where the cross attention is calculated following Equ. \ref{equ:ca}. 

\textbf{Mutation prediction heads.} After reconstructing the mutant representation by $h_{\text{mt}}=h_{\text{wt}}+h_\Delta$, we develop a position prediction head $f_{\text{pos}}$ and a language modeling head $f_{\text{LM}}$ to predict the mutation. Specifically:
\begin{equation}
\begin{aligned}
& P\left(x_i^{(\text{mt})}\neq x_i^{(\text{wt})}\right)=f_{\text{pos}}\left(h_i^{\text{(mt)}}\right),\\
& P\left(x_i^{(\text{mt})}\right)=f_{\text{LM}}\left(h_i^{(\text{mt})}\right),\\
\end{aligned}
\end{equation}
where $P\left(x_i^{(\text{mt})}\neq x_i^{(\text{wt})}\right)$ denotes the probability of $i$-th amino acid to be mutated, and $P\left(x_i^{(\text{mt})}\right)$ denotes the probability distribution of the $i$-th amino acid. The parameters of the position prediction head are initialized from scratch, and those of the language modeling head are derived from the PLM.

\subsubsection{Justifications for Mutational Features}
To model mutations explicitly, we leverage the subtraction of the wild-type and mutant representations as the mutational features $h_{\Delta}$, which is subsequently processed by the \textit{delta} encoder. One of the essential considerations is that the PLM is overly smooth, making $h_{\Delta}$ too small and less informative. However, we argue that due to the non-smooth nature of the protein fitness landscape \cite{kirjner2023optimizing}, the output representations of PLMs are also non-smooth. Moreover, after training, the \textit{delta} encoder learns to capture the orientation of $h_{\Delta}$, yielding a $z_{\Delta}$ with an appropriate norm. We also present empirical justification by calculating the average $l_2$-norm of $h_{\text{wt}}$, $h_{\Delta}$, and $z_{\Delta}$ on MutaDescribe, which are displayed in Tab. \ref{tab:norm}.

\begin{table}[tpb]
\centering
\captionof{table}{\textbf{Average $l_2$-norm of MutaPLM's intermediate representations on MutaDescribe.}}
\label{tab:norm}
\begin{tabular}{lccc}
\toprule
Representation & $h_{\text{wt}}$   &  $h_{\Delta}$     & $z_{\Delta}$     \\
\midrule
Avg. $l_2$-norm        & 9.90 & 0.35 & 1.04 \\
\bottomrule
\end{tabular}
\end{table}

\subsubsection{Pre-training Objectives \label{app:train}}
MutaPLM performs pre-training on large-scale protein-relevant literature. Given the protein sequence $x_{\text{wt}}$ and its semantically related text $t$, we optimize the following objectives:

\textbf{Protein-to-text generation.} We first concatenate the latent wild-type features $z_{\text{wt}}$ in Equ. \ref{equ:ca} and the text embeddings $e$ in Equ. \ref{equ:text}. We perform conditional auto-regressive language modeling that aims to generate $t$ based on the protein representations and previous tokens. The objective is calculated as follows:
\begin{equation}
\begin{aligned}
    &z=[\underbrace{z_1,z_2,\cdots,z_K}_{\text{protein}},\underbrace{z_{K+1},\cdots,z_{K+N}}_{\text{text}}]=g_{\text{transformers}}\left([z_{\text{wt}};e]\right),\\
    &P(t_i|t_{<i},z_{\text{wt}})=g_{\text{LM}}(z_{K+i}),\\
    &\mathcal{L}_{p2t}=\frac{1}{N}\sum_{i=1}^{N}H\left[t_i, P(t_i|t_{<i},z_{\text{wt}})\right],
\end{aligned}
\end{equation}
where $H(\cdot,\cdot)$ denotes cross-entropy. 

\textbf{Text-to-protein generation.} We first append $K$ trainable soft tokens $s=[s_1,s_2,\cdots,s_K]$ to the input token embeddings to summarize textual semantics. Then, we derive $z_\Delta$ as the last hidden state of $s$ as follows:
\begin{equation}
    \tilde{z}=[\underbrace{\tilde{z}_1,\tilde{z}_2,\cdots,\tilde{z}_N}_{\text{text}},\underbrace{\tilde{z}_{N+1},\cdots,\tilde{z}_{N+K}}_{z_{\Delta}}]=g_{\text{transformers}}([e;s]),
\end{equation}
where $s$ denotes the soft tokens. We pass $z_{\Delta}$ into the \textit{delta} decoder to obtain $h_{\Delta}$ as in Equ. \ref{equ:delta_dec}. It is worth noting that in this stage, we are aimed at aligning the feature space of PLMs and LLMs, and $z_{\Delta}$ and $h_{\Delta}$ are \textbf{NOT} related to protein mutations.

Then, we randomly mask 15\% amino acids in the protein sequence. We adopt the conditional masked language modeling objective to reconstruct the masked tokens as follows:

\begin{equation}
\begin{aligned}
&h_{\text{mask}}=f_{\text{PLM}}(x_{\text{mask}}),\\
&\tilde{h}=\left[\tilde{h}_1,\tilde{h}_2,\cdots,\tilde{h}_L\right]=h_{\text{mask}}+h_{\Delta},\\
&P\left(x_i^{(\text{wt})}\big|x_{\text{mask}},h_{\Delta}\right)=f_{\text{LM}}\left(\tilde{h}_i\right),\\
&\mathcal{L}_{t2p}=\frac{1}{|\mathcal{M}|}\sum_{i\in \mathcal{M}}H\left[x_i,P\left(x_i^{(\text{wt})}\big|x_{\text{mask}},h_{\Delta}\right)\right],
\end{aligned}
\end{equation}
where $x_{\text{mask}}$ is the masked sequence of the wild-type $x_{\text{wt}}$, and $\mathcal{M}$ denotes the masked positions.

\textbf{Overall objective.} The overall objective for pre-training is calculated by:
\begin{equation}
    \mathcal{L}_1=\mathbb{E}_{(x_{\text{wt}},t)\sim \mathcal{D}_1}(\mathcal{L}_{p2t}+\mathcal{L}_{t2p}),
\end{equation}
where $\mathbb{E}$ denotes expectation, and $\mathcal{D}_1$ denotes our pre-training dataset.
\subsubsection{Fine-tuning Objectives}
\begin{table}[tpb]
\small
\captionsetup{font={small,stretch=0.95}}
\caption{\textbf{Prompt templates for fine-tuning.} The first and second round dialogs are composed of system prompts, latent wild-type and \textit{delta} features, and special tokens including \texttt{<BOP>, <EOP>, <BOM>, <EOM>}. We highlight the parts that are used to calculate the objectives.} 
\label{tab:prompt_finetune}
\centering
\begin{tabular}{p{3.5cm}p{10cm}}
\toprule
Type                           & Content \\
\midrule
System Prompt                 & You are an expert at biology and life science. Now a user gives you several protein sequences and mutations. Please follow user instructions and answer their questions.        \\
\midrule
User Prompt for Function Prediction           & Based on the following protein sequence, please describe its function.        \\
\midrule
User Prompt for Mutation Explanation       & Next is a mutation from <$x_i$> to <$\hat{x}_i$> at position $i$. Please generate a brief/detailed introduction to describe it.        \\
\midrule
User Prompt for Mutation Engineering        & Next is a brief/detailed introduction of mutational effects. Please generate a protein mutation that fits the description.        \\
\midrule
Round 1 Dialog                 & \texttt{[System Prompt] [User Prompt for Function Prediction] <BOP> [Latent Wild-type Features] <EOP> \textcolor{red}{[Protein Function]}}        \\
\midrule
Round 2 Dialog for Mutation Explanation &  \texttt{[Round 1 Dialog] [User Prompt for Mutation Explanation] <BOM> [}\textit{Delta} \texttt{Features] <EOM> \textcolor{red}{[Mutational Effects]}}       \\
\midrule
Round 2 Dialog for Mutation Engineering &  \texttt{[Round 1 Dialog] [User Prompt for Mutation Engineering] [Mutational Effects] <BOM> \textcolor{red}{[Soft Embeds]} <EOM>}       \\
\bottomrule
\end{tabular}
\end{table}

We employ a chain-of-thought (CoT) strategy to reason over protein functions and mutational effects in a two-round dialog. Given the wild type sequence $x_{\text{wt}}$, the mutant sequence $x_{\text{mt}}$, the description of protein functions $t_{\text{func}}$ and the description of mutation effects $t_{\Delta}$, we calculate the following objectives:

\textbf{First-round dialog.} We first prompt the LLM to generate function descriptions $t_{\text{func}}=\left[t_1^{\text{(func)}}, t_2^{\text{(func)}},\cdots,t_M^{\text{(func)}}\right]$ based on the wild-type protein. We perform conditional auto-regressive language modeling as follows:
\begin{equation}
    \mathcal{L}_{\text{func}}=\frac{1}{M}\sum_{i=1}^{M}H\left[t_i^{\text{(func)}},P\left(t_i^{\text{(func)}}\big|t_{<i}^{\text{(func)}},z_{\text{wt}}\right)\right].
\end{equation}

The predictions of protein functions $y_{\text{func}}=\left[y_1^{\text{(func)}},y_2^{\text{(func)}},\cdots,y_N^{\text{(func)}}\right]$ is derived by:
\begin{equation} y_i^{\text{(func)}}=\text{argmax}\left\{P\left(y_i^{\text{(func)}}\big|y_{<i}^{\text{(func)}},z_{\text{wt}}\right)\right\}
\end{equation}
\textbf{Second-round dialog for mutation explanation.} We prompt the LLM to generate textual descriptions for mutation effects $t_{\Delta}=\left[t_1^{(\Delta)}, t_2^{(\Delta)},\cdots,t_T^{(\Delta)}\right]$ based on the function information in the first-round dialog and protein \textit{delta} features $z_\Delta$. The objective is calculated as follows:
\begin{equation}
    \mathcal{L}_{\text{exp}}=\frac{1}{T}\sum_{i=1}^{T}H\left[t_i^{(\Delta)},P\left(t_i^{(\Delta)}\big|t_{<i}^{(\Delta)},y_{\text{func}},z_{\Delta},z_{\text{wt}}\right)\right].
\end{equation}

\textbf{Second-round dialog for mutation engineering.} We apply the same soft tokens $s$ as in pre-training to the input prompt to calculate the \textit{delta} features based on the first-round dialog and descriptions of mutational effects:
\begin{equation}
    \hat{z}=[\underbrace{\hat{z}_1,\hat{z}_2,\cdots,\hat{z}_N}_{\text{prompt}},\underbrace{\hat{z}_{N+1},\cdots,\hat{z}_{N+K}}_{z_{\Delta}}]=g_{\text{transformers}}([t_{\text{prompt}};s]),
\end{equation}
where $t_{\text{prompt}}$ is the input embeddings of the prompt involving the first-round dialog and the mutational effects.

Then, reconstructing $h_{\text{mt}}=h_{\text{wt}}+h_{\Delta}$ with the \textit{delta} decoder, we calculate the weighted cross-entropy loss for the mutation position and the mutated amino acid with the prediction heads:
\begin{equation}
\begin{aligned}
    \mathcal{L}_{\text{eng}}=-\frac{1}{L}\sum_{i=1}^{L}&\left\{\mathbbm{1}\left\{x_i^{(\text{mt})}=x_i^{(\text{wt})}\right\}\log (1-f_{\text{pos}}(h_i^{\text{mt}}))\right.\\&+\lambda\cdot\mathbbm{1}\left\{x_i^{(\text{mt})}\neq x_i^{(\text{wt})}\right\}\log f_{\text{pos}}(h_i^{\text{mt}})\\&\left.-L\cdot \mathbbm{1}\left\{x_i^{(\text{mt})}\neq x_i^{(\text{wt})}\right\}H\left[x_i^{\text{(mt)}},f_{\text{LM}}(h_i^{\text{(mt)}})\right]\right\},
\end{aligned}
\end{equation}
where $\mathbbm{1}\{\cdot\}$ is the boolean indicator function, and $\lambda$ is a hyper-parameter controlling label weight. In our experiments, we set $\lambda=50$.  

The overall objective is calculated as follows:
\begin{equation}
    \mathcal{L}_2=\mathbb{E}_{(x_{\text{wt}},x_{\text{mt}},t_{\text{func}},t_\Delta)\sim \mathcal{D}_2}(\mathcal{L}_{\text{func}}+\mathcal{L}_{\text{exp}}+\mathcal{L}_{\text{eng}}),
\end{equation}
where $\mathcal{D}_2$ is our fine-tuning dataset.

The prompt templates for fine-tuning are displayed in Tab. \ref{tab:prompt_finetune}.

\subsection{Training data \label{app:data}}
\subsubsection{Pre-training Data\label{app:pt_data}}
Our pre-training data involves 1.1M protein-text pairs collected from the UniProtKB/SwissProt \cite{boutet2016uniprotkb} database. We download 467.8K proteins with the \textit{Publications} entry and retrieve 257.2K PubMed \cite{canese2013pubmed} abstracts based on the reference information.

\subsubsection{Fine-tuning and Testing Data: MutaDescribe \label{app:MutDes-description}}

To create a natural language annotated dataset for protein mutations, we first collect 164K samples from the \textit{Phenotypes \& Variants} entry of UniProtKB/SwissProt. After deduplication and removing sites without valid text annotations, we obtain 107K mutants for 21K proteins as our raw data, comprising 33K natural variants and 74K mutagenesis sequences. 



\begin{table}[tpb]
\captionsetup{font={small,stretch=0.95}}
    \centering
    \caption{\textbf{An Overview of MutaDescribe.}}
    \begin{tabular}{cccc}
        \toprule
        \# All & \# Raw & \# Enriched & \# Reversed \\
        \midrule
        171,147 & 106,645 & 57,147 & 64,502 \\
        \midrule
        \multicolumn{2}{c}{Avg. words (UniProtKB)} & \multicolumn{2}{c}{Avg. words (Enriched)}\\
        \midrule
        \multicolumn{2}{c}{9.44} & \multicolumn{2}{c}{28.33}\\
        \midrule
        \# Malignant & \# Benign & \# Not significant & \# Unknown \\
        \midrule
        72,198 & 8,000 & 26,447 & 4\\
        \bottomrule
    \end{tabular}
    \label{tab:mutdes-detail}
\end{table}

Unfortunately, the collected raw data is not suitable for protein mutation modeling, mainly owing to the following problems: (1) As shown in Tab. \ref{tab:mutdes-detail}, the expert-revised annotations within UniProtKB contain an average of 9.4 words, containing limited information. (2) Through analyzing the polarity of the mutational effects, we observe that the number of malignant and benign mutations are imbalanced ($\sim$ 9:1), which may mislead model predictions. 

To address these issues, (1) we perform data enrichment by collecting the abstracts of the biological literature in which the mutation is mentioned. We retrieve 50K publications based on the reference information of the mutation available in UniProtKB and prompt GPT-3.5-turbo to extract relevant information from the abstracts. The prompt template is visualized in Tab. \ref{tab:app-prompt-enrich}. After ChatGPT enrichment, the textual annotations are expanded with an average of 28.3 words. (2) We generate 64.5K additional reverse samples. Specifically, for each malignant and benign mutation, we exchange the wild-type and mutant and prompt GPT-3.5-turbo to flip the polarity of the textual descriptions for mutational effects. We empirically find that the quality of mutation descriptions using GPT-3.5-turbo and GPT-4 is similar, and therefore we opt for GPT-3.5-turbo to save API costs.


We implement two splitting strategies for our dataset. For \textbf{structural split}, we first partition our dataset into training, validation, and test sets. Then, for each wild-type sequence in the test set, we calculate the maximum sequence homology with the wild-type sequences in the training set by MMseqs2 \cite{steinegger2017mmseqs2}. Based on the homology, we divide the test set into three subsets. The \textit{Easy} subset comprises 460 mutants with homology between 0.95 and 1, the \textit{Medium} subset comprises 384 mutants with homology between 0.5 and 0.95, and the \textit{Hard} subset comprises 404 mutants with homology between 0 and 0.5. For \textbf{temporal split}, we extract the publication date of the literature reporting each mutation. Mutations studied before 2022 are used as training and validation sets, while those studied in 2022 and 2023 comprise the test set. The train/valid/test set comprises 156K, 8K, and 1.6K samples, respectively. The detailed statistics of temporal split are shown in Tab. \ref{tab:app-temporal-dataset}.

\begin{table}[pb]
\captionsetup{font={small,stretch=0.95}}
\captionof{table}{\textbf{Statistics of the temporal split.} We report the number of proteins and samples, the average protein sequence length, and the average number of words for mutational effects.}
\label{tab:app-temporal-dataset}
\centering
\begin{tabular}{lcccc}
\toprule
Split   & \# Proteins & \# Samples & Avg. sequence length & Avg. words \\
\midrule
Train  & 20,295 & 156,300 & 518.00 & 28.48 \\
Valid  & 5,436 & 8,000 & 514.30 & 28.73 \\
Test   & 310 & 1,611 & 536.67 & 26.37 \\
\bottomrule
\end{tabular}
\end{table}

\begin{figure}[tpb]
\captionsetup{font={small,stretch=0.95}}
    \centering
    \includegraphics[width=\linewidth]{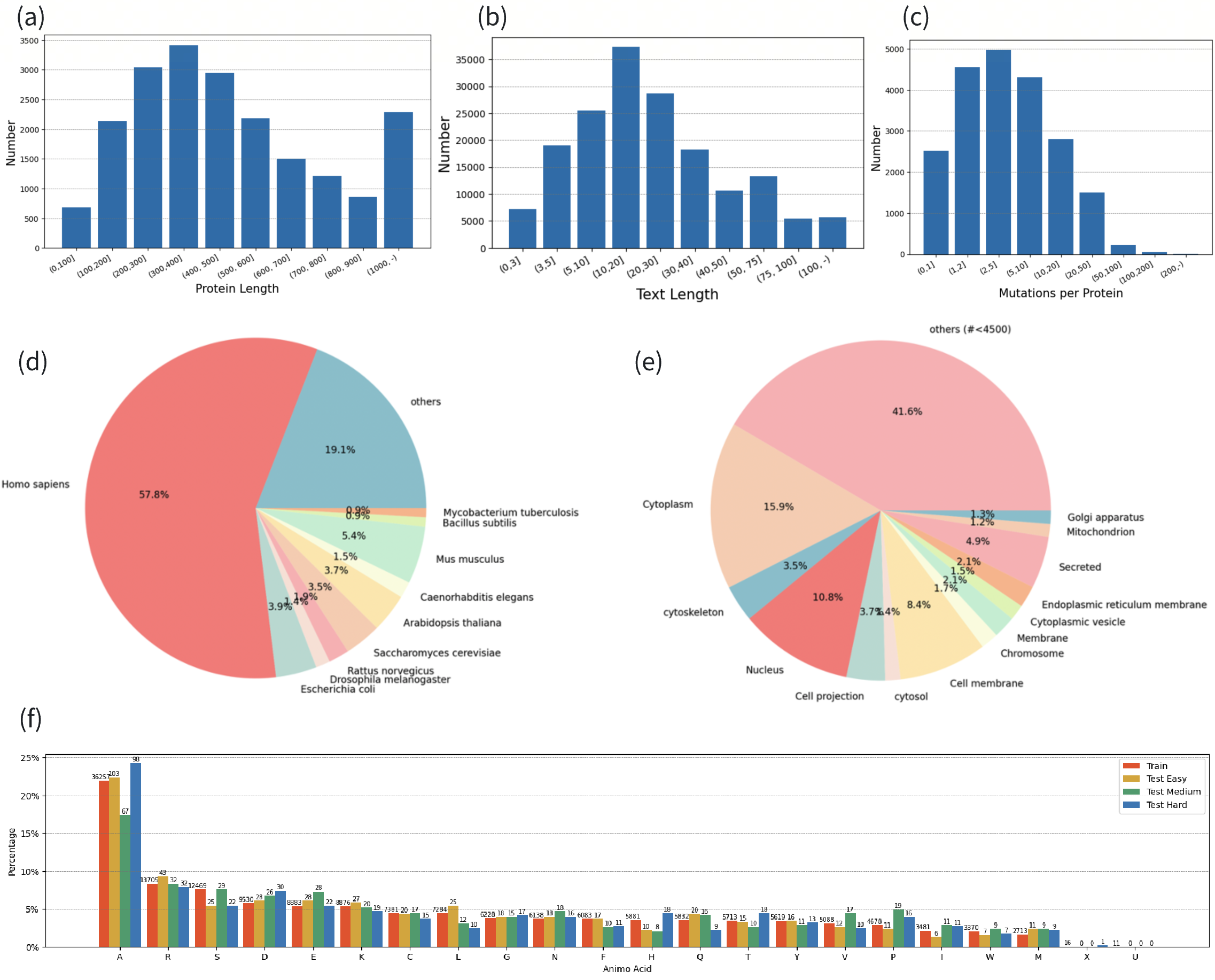}
    \caption{\textbf{Detailed statistics of the MutaDescribe dataset.} We show (a) the length of protein sequences, (b) the number of words in textual annotations, (c) the number of mutation samples per protein, (d) the distribution of the originating species, (e) the distribution of the cellular localization and (f) the distribution of the mutated amino acid.}
    \label{fig:app-dataset}
\end{figure}

We present a closer look at our MutaDescribe dataset in Fig. \ref{fig:app-dataset}, displaying the length of protein sequences, the number of words in textual annotations, the number of mutation samples per protein, the distribution of the originating species, the distribution of the cellular localization and the distribution of the mutated amino acid. We show in our illustrations that MutaDescribe is a large-scale, diverse, and detailed annotated dataset for studying protein mutations.

\begin{table}[tpb]
\captionsetup{font={small,stretch=0.95}}
\centering
\caption{\textbf{Prompt template for data enrichment.} We prompt GPT-3.5-turbo to extract relevant information from the abstracts of the biological literature in which the mutation is mentioned.}
\begin{tabular}{p{13.5cm}}
    \toprule
    \texttt{[System prompt]}You will be provided with a document and some relevant mutation sites (for example, site A21D indicates a mutation from A to D at position 21). First, determine whether these sites are mentioned in the document. If so, extract the text from the document that describes the functional changes caused by these sites. Otherwise, you must extract any functional changes mentioned in the document. For each site, please try to extract the corresponding protein name or gene name. You must be accurate and clear. Return a series of JSON documents, with each JSON formatted as follows:\newline
    \{"Mutation Site": <provided mutation site>,\newline
    "Mentioned": <whether this site is mentioned in the document>\newline
    "protein\_name": <protein name corresponding to the site>,\newline
    "gene\_name": <gene name corresponding to the site>,\newline
    "Functional\_changes": <functional information>\}\newline
    \newline
    \texttt{[User prompt]} \newline
    document: \texttt{<document>}\newline
    sites: \texttt{<list of mutation sites>}
    \\
    \bottomrule
\end{tabular}
\label{tab:app-prompt-enrich}
\end{table}

\subsection{Experiment Settings \label{app:exp}}

\begin{table}[htpb]
\small
\captionsetup{font={small,stretch=0.95}}
\caption{\textbf{Prompt templates for each baseline for mutation explanation.} \texttt{\{original\_aa\}} and \texttt{\{mutated\_aa\}} denote the amino acid before and after the mutation respectively. \texttt{\{fitness\_change\}} is the subtraction of the PLM-calculated evolutionary plausibility scores between the mutant and wild-type.}
\centering
\begin{tabular}{p{4cm}p{9cm}}
    \toprule
    Baseline                    &       Prompt \\
    \midrule
    Galactica-6.7B              &   protein \texttt{\{protein\_name\}}: \texttt{[START\_AMINO] \{protein\_sequence\} [END\_AMINO]} has a mutation \texttt{\{original\_aa\}} to \texttt{\{mutated\_aa\}} at position \texttt{\{position\}}. Question: What are the functional changes of the protein after this mutation?\newline
    Answer:\\
    \midrule
    ProLLaMA                    &  Seq=<\texttt{\{protein\_sequence\}}> has a mutation \texttt{\{original\_aa\}} to \texttt{\{mutated\_aa\}} at position \texttt{\{position\}}. Question: What are the functional changes of the protein after this mutation? \newline
    Answer:\\
    \midrule
    Mol-Instructions            & Please evaluate protein \texttt{\{protein\_name\}} with the given mutation, and provide an explanation of any activity or reaction the mutation may cause: \newline
    <protein> ```\texttt{\{protein\_sequence\}}```\newline
    <mutation> \texttt{\{original\_aa\}} to \texttt{\{mutated\_aa\}} at position \texttt{\{position\}} \\
    \midrule
    GPT-4-0613 (few-shot)          & \texttt{[System prompt]} You are an expert in bioinformatics. You will be provided with a protein and its mutation information. Please predict the changes in the protein's function after this mutation. Your response should only focus on the effect of the change without additional words. \newline
    \newline
    \texttt{[User prompt]} Example 1:\newline
    protein name: Glutathione S-transferase P\newline
    protein sequence: MPPYTVVYFPVRGRCAALRM...\newline
    mutation: D to A at position 99\newline
    \texttt{(Additional samples ...)} \newline
    \newline
    protein name: \texttt{\{protein\_name\}}\newline
    protein sequence: \texttt{\{protein\_sequence\}}\newline
    mutation: \texttt{\{original\_aa\}} to \texttt{\{mutated\_aa\}} at position \texttt{\{position\}}\newline
    function change: \\
    \midrule
    GPT-4 + ESM-2 \& \newline
    GPT-4 + OntoProtein           &     \texttt{[System prompt]} You are an expert in bioinformatics. You will be provided with a protein and its fitness score after a single mutation. Please predict the changes in the protein's function based on the fitness score. Your response should only focus on the effect of the change without additional words. \newline
    \newline
    \texttt{[User prompt]} Example 1:\newline
    protein name: Glutathione S-transferase P\newline
    protein sequence: MPPYTVVYFPVRGRCAALRM...\newline
    mutation: D to A at position 99\newline
    fitness change: -0.7684 \newline
    \texttt{(Additional samples ...)} \newline
    \newline
    protein name: \texttt{\{protein\_name\}}\newline
    protein sequence: \texttt{\{protein\_sequence\}}\newline
    mutation: \texttt{\{original\_aa\}} to \texttt{\{mutated\_aa\}} at position \texttt{\{position\}}\newline
    fitness change: \texttt{\{fitness\_change\}} \newline
    function change:          \\
    \bottomrule
\end{tabular}
\label{tab:explain-baseline-prompt}
\end{table}

\subsubsection{Baselines for Mutation Explanation \label{app: baseline_setup}}
For mutation explanation, we implement the following baselines:

\textbf{Galactica-6.7B \cite{taylor2022galactica}.} This baseline is a unified large-language model pre-trained on scientific papers and protein knowledge bases. We prompt the model to investigate if it could explain mutational effects in a zero-shot manner.

\textbf{ProLLaMA \cite{lv2024prollama}.} This baseline is developed on LLaMA2-7B by further pre-training the model on protein sequences from UniRef50 \cite{suzek2015uniref}. Similarly, we perform zero-shot mutation explanation by prompting.

\textbf{Mol-Instructions \cite{fang2023mol}.} We implement the protein-oriented model of Mol-Instructions that is instruction-tuned from LLaMA2-7B \cite{touvron2023llama}. We perform zero-shot prompting that provides the model with the name and amino acid sequence of the protein sequence and task definitions.

\textbf{GPT-4 \cite{achiam2023gpt} with in-context learning.} We adopt the 0613 version of GPT-4, the most advanced LLM in natural language processing. In addition to the protein name, wild-type sequence, and mutation information, we provide few-shot demonstrations to facilitate in-context learning. For the 1-shot and 5-shot baseline, we randomly sample 1 and 5 samples from the training set of MutaDescribe. For the kNN-based 5-shot baseline, we follow \cite{ai4science2023impact} to search for relevant samples based on the sequence homology calculated by MMseqs2 \cite{steinegger2017mmseqs2}. We select 5 samples from the training set with the highest homology as few-shot demonstrations for each test sample.

\textbf{GPT-4 + ESM-2 \cite{lin2023evolutionary}.} ESM-2 is a popular protein language model pre-trained on evolutionary-scale databases. Given a mutation, we mask the mutated position and utilize ESM-2 (650M) to predict the logits for the mutated amino acid. Following \cite{meier2021language}, we adopt the subtraction between the mutant and wild-type logits as the evolutionary plausibility scores. We follow the 5-shot kNN setting on GPT-4 and provide the scores as additional information.

\textbf{GPT-4 + OntoProtein \cite{zhang2021ontoprotein}.} OntoProtein is a text-augmented PLM that aligns protein sequences with gene ontology definitions. We follow the GPT-4 + ESM-2 baseline to predict mutational effects based on evolutionary plausibility and kNN few-shot demonstrations.

\textbf{AugmentedESM \cite{hsu2022learning}.} In the original paper, the model is designed to solve fitness regression tasks by linearly combining the adaptive fitness score calculated following \cite{meier2021language} and the amino acid sequence. We slightly adapt the model to perform mutation explanation by feeding the fitness score and the raw protein sequence into BioMedGPT-LM. We fine-tune the LLM with the casual auto-regressive language modeling objective on mutation effects. The hyperparameters for fine-tuning are the same as MutaPLM.

\textbf{Finetuned ESM-2.} Similar to MiniGPT-4 \cite{zhu2023minigpt}, we translate each residue representation of ESM-2 (650M) \cite{lin2023evolutionary} into LLM input embeddings using a linear projection layer. We fine-tune BioMedGPT-LM with the casual auto-regressive language modeling objective on mutation effects based on the translated features of the wild-type and mutant. The hyperparameters for fine-tuning are also the same as MutaPLM.

The prompts for our baselines are displayed in Tab. \ref{tab:explain-baseline-prompt}.

\subsubsection{Baselines for Mutation Engineering \label{app:baseline_eng}}
For mutation engineering, we implement the following baselines:

\textbf{Random.} As the name suggests, the proposed mutations are randomly sampled from every possible single-site substitution with equal probability.

\textbf{GPT-4 \cite{achiam2023gpt} with in-context learning.} We provide few-shot examples for GPT-4 to suggest protein mutations, and the sampling strategy is the same as in mutation explanation. We evaluate accuracy and top-50 recall with a two-round dialog. In the first-round dialog, we directly prompt GPT-4 to provide 50 mutations on arbitrary positions. In the second-round dialog, we provide the model with the ground-truth position and ask 

\textbf{ESM-2 \cite{lin2023evolutionary}.} We feed the whole sequence into the PLM to calculate the output logits for each amino acid. We rank mutations by the subtraction of the mutant and wild-type logits.

\textbf{OntoProtein \cite{zhang2021ontoprotein}.} This baseline follows the same implementation as ESM2-650M.

\textbf{ProtST (ESM-2) \cite{xu2023protst}.} ProtST trains a series of PLMs by contrastive learning \cite{radford2021learning} between protein sequences and biomedical texts. Hence, we implement a cross-modal retrieval strategy, using the cosine similarity between the mutated sequence and the textual description of mutational effects to score mutations. We opt not to report top-50 recall scores due to: (1) unaffordable computational costs, as each possible mutation requires an individual forward pass, and (2) poor performance, as the baseline merely outperforms random guesses. 

\textbf{Fine-tuned BioMedGPT.} We provide the LLM with the wild-type sequence and textual instructions of desired mutational effects, and fine-tune the model to propose mutations. To evaluate accuracy, we additionally provide the mutated position and prompt the model to generate the mutated amino acid. To evaluate top-50 recall, we prompt the model to generate a single mutation, since our dataset only comprises one ground-truth mutation. The evaluations are performed within two independent sessions, and we combine the causal auto-regressive language modeling objective of both sessions during fine-tuning.

\textbf{Fine-tuned ESM-2.} We leverage BioMedBERT \cite{gu2021domain} to encode the textual instructions. We employ a cross-attention layer that takes the ESM-2 representations of the wild-type sequence as queries and the BioMedBERT representations as keys and values. The outputs are fed into a position prediction head and a language modeling head to predict mutations, which is the same as MutaPLM.

The prompt templates for GPT-4 and fine-tuned BioMedGPT are presented in Tab. \ref{tab:app-prompt-engineering}.

\begin{table}[tpb]
\captionsetup{font={small,stretch=0.95}}
\centering
\caption{\textbf{Prompt template for few-shot GPT-4 and fine-tuned BioMedGPT in mutation engineering.} }
\begin{tabular}{p{3cm}p{10.5cm}}
    \toprule
    Evaluating Rec@50 on GPT-4    &
    \texttt{[System prompt]} 
    You are an expert in bioinformatics. You will be provided with a protein and the functional change resulting from a single-site mutation. Please predict the 50 most probable mutation sites where Each entry starts with the amino acid before the mutation, followed by the position of the mutation, and ends with the amino acid after the mutation. For example, D65A indicates that the amino acid at position 65 changes from D to A. Your response should only contain the 50 sites in a list format separated by commas, without additional words. \newline
    \newline
    \texttt{[User prompt]} Example 1:\newline
    protein name: Glutathione S-transferase P\newline
    sequence: MPPYTVVYFPVRGRCAALRMLLA...\newline
    functional change: Reduces affinity for glutathione.\newline
    50 probable mutation sites: D99A, T110K, D58V, L53I, V165P, ...\newline
    \texttt{(Additional samples ...)}\newline
    \newline
    protein name: \texttt{\{protein name\}}\newline
    sequence: \texttt{\{protein sequence\}}\newline
    functional change: \texttt{\{mutational effects\}}\newline
    50 probable mutation sites: \\

    \midrule
    Evaluating Accuracy on GPT-4   &
    \texttt{\{First round dialog\}}\newline
    \texttt{[User prompt]} The correct mutated position is \texttt{\{mutation position\}}. What is the most probable amino acid after the mutation? The valid amino acids include: [G, V, S, E, C, K, Q, N, M, H, I, Y, L, D, W, A, T, R, P, F]. Your answer should only contain one of the uppercase amino acids without other words. 
    \\
    \midrule
    Evaluating Rec@50 on fine-tuned BioMedGPT & You are an expert assistant in biology and protein engineering. Now you are given a protein sequence and an instruction describing a mutation effect. \newline 
    \newline 
    Protein: \texttt{\{protein sequence\}}\newline
    Instruction: \texttt{\{mutational effects\}}\newline
    User: Please design a mutation that best fits the instruction.\newline
    Assistant:
    \\
    \midrule
    Evaluating Accuracy on fine-tuned BioMedGPT & You are an expert assistant in biology and protein engineering. Now you are given a protein sequence and an instruction describing a mutation effect. \newline 
    \newline 
    Protein: \texttt{\{protein sequence\}}\newline
    Instruction: \texttt{\{mutational effects\}}\newline
    User: Given mutation at position \texttt{\{mutation position\}}, please choose an amino acid that best fits the instruction.\newline
    Assistant:
    \\
    \bottomrule
\end{tabular}
\label{tab:app-prompt-engineering}
\end{table}

\subsubsection{Human-AI Collaborative Evaluation for Mutation Explanation \label{app: prompt_eva}}
Due to the complexity of biomedical texts, we develop a human-AI collaborative evaluation pipeline to comment on the accuracy and helpfulness of predicted mutational effects. Specifically, we query GPT-4 to compare model predictions with ground-truth annotations as in Tab. \ref{tab:app-prompt-eva} and categorize them as follows. 

\begin{itemize}[noitemsep,topsep=0pt,parsep=4pt,partopsep=0pt,leftmargin=25pt]
    \item \textit{Accurate.} The predicted alterations in protein functions and estimations of extent are mostly the same as the ground truth.
    \item \textit{Relevant.} The prediction identifies the protein function that is altered by the mutation. While it accurately predicts the attenuation or the degradation, the estimation of the extent is not correct.
    \item \textit{Opposite.} The prediction identifies the protein function that is altered by the mutation. However, it mistakenly predicts attenuation as degradation or vice versa.
    \item \textit{Irrelevant.} The prediction and the ground truth are about completely different functional alterations.
\end{itemize}

\begin{figure}[tpb]
    \centering
    \captionsetup{font={small,stretch=0.95}}
    \includegraphics[width=0.6\linewidth]{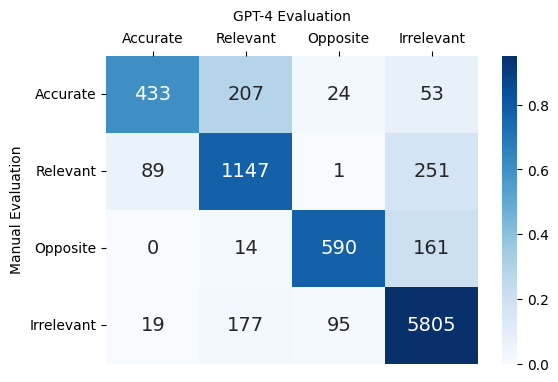}
    \caption{\textbf{Confusion matrix between GPT-4 and manual evaluation.}}
    \label{fig:app-confusion}
\end{figure}

Then, we recruit a postgraduate from a top university who majors in biology to further assess the results. Specifically, we collect samples that are marked as \textit{Accurate}, \textit{Relevant}, and \textit{Opposite} by GPT-4, and include \textit{Irrelevant} samples for strong baselines (5-shot GPT-4 models and fine-tuned models) and MutaPLM. We present the mutation explanations, ground-truth results, GPT-4 evaluation, and categorization protocol, and ask the expert to rectify the evaluation result if necessary. In total, 12.0\% of the GPT-4 evaluations are modified, and the confusion matrix is displayed in Fig. \ref{fig:app-confusion}. We observe that GPT-4 evaluation is consistent with human experts in most cases, showcasing its reliability as a proxy of expert evaluators in saving evaluation costs. However, it occasionally misclassifies \textit{Accurate} predictions into \textit{Relevant}, and \textit{Relevant} or \textit{Opposite} predictions into \textit{Irrelevant}, which we attribute to the fact that GPT-4 tends to favor more fluent answers instead of more informative ones. We leave more realistic and labor-saving evaluation strategies for future exploration.

\begin{table}[tpb]
\captionsetup{font={small,stretch=0.95}}
\centering
\caption{\textbf{Prompt template for GPT-4 evaluation.} We leverage GPT-4 to categorize predictions into \textit{Accurate, Relevant, Opposite}, and \textit{Irrelevant}, based on the relevance between the predicted functional alterations and ground-truth explanations.}
\begin{tabular}{p{13.5cm}}
    \toprule
    \texttt{[System prompt]} You are an expert in biology and protein sciences. You want to figure out the effects of protein mutations by alterations of protein functions. Now we provide you with two descriptions of protein mutational effects in a JSON format, where the "label" denotes the ground truth description of the mutational effects, and the "prediction" denotes the prediction of a model. You should be precise and faithful in evaluating if the predicted mutation effects are semantically related to the ground truth. You should answer with one of the following categories:
    \newline\newline
    (1) Accurate. The prediction and the label describe the same functions that are altered, and the extent of functional changes is mostly the same (For example, "strongly decrease" and "abolish").\newline
    (2) Relevant. The prediction and the label describe the same functions that are altered, and the extent of functional changes are in the same direction (For example, "strongly increase" and "slightly increase").\newline
    (3) Opposite. The prediction and the label describe the same functions that are altered, but the functional changes are opposite (For example, "increase" and "decrease").\newline
    (4) Irrelevant. The prediction and the label describe different alterations of functions.\newline
    Note that you should be careful about the altered functions before analyzing the extent. Answer with one word only from "Accurate", "Relevant", "Opposite" and "Irrelevant" to summarize your evaluation.
    \newline \\
   \texttt{[User prompt]}\{"label": \texttt{\{ground\_truth\}}, "prediction": \texttt{\{model\_output\}}\} \\
    \bottomrule
\end{tabular}
\label{tab:app-prompt-eva}
\end{table}

\subsubsection{Multi-round Optimization \label{app:optimize}}
We incorporate the following datasets from \cite{ren2022proximal} for multi-round fitness optimization:
\begin{itemize}[noitemsep,topsep=0pt,parsep=4pt,partopsep=0pt,leftmargin=25pt]
    \item \textbf{Adeno-associated Viruses (AAV)} \cite{bryant2021deep}. The dataset involves a 28-amino acid segment of the \textit{caspid protein VP1} from \textit{Adeno-associated virus}. The optimization objective is to improve its capability as a gene delivery vector.
   \item \textbf{Aliphatic Amide Hydrolase (AMIE)} \cite{wrenbeck2017single}. The dataset aims to improve the enzymic activity of \textit{Aliphatic amidase} from \textit{Pseudomonas aeruginosa} in catalyzing the hydrolysis of short-chain aliphatic amides.
   \item \textbf{Green Fluorescent Proteins (avGFP)} \cite{sarkisyan2016local}. The dataset aims to enhance the fluorescent intensity of the \textit{Green Fluorescent Protein} from \textit{Aequorea victoria}. The protein is widely adopted as a biosensor for detecting gene expressions and protein locations.
   \item \textbf{Ubiquitination Factor Ube4b (E4B)} \cite{starita2013activity}. The dataset aims to improve the enzymic activity of \textit{Ubiquitin conjugation factor E4B} in \textit{Homo sapiens}, which plays a role in proteasomal degradation by interacting with other proteins.
   \item \textbf{Levoglucosan Kinase (LGK)} \cite{klesmith2015comprehensive}. The dataset focuses on \textit{Levoglucosan kinase} in \textit{Lipomyces starkeyi}. The optimization objective is to enhance its catalytic activity in canonical kinase phosphotransfer reaction.
   \item \textbf{SUMO E2 conjugase (UBE2I) \cite{weile2017framework}.} The dataset studies \textit{SUMO-conjugating enzyme UBC9} in \textit{Homo sapiens} which is relevant to several human diseases. The optimization objective is to improve the growth rescue rate at high temperatures in a yeast strain.
\end{itemize}

We manually write prompts in Tab. \ref{tab:fitopt} to navigate the optimization process by a beam search process. Specifically, we initialize the candidate set with the wild-type sequence. Then, for each round of optimization, we feed each candidate sequence and the textual instruction into the decoding workflow of MutaPLM. Then we sample $K$ mutations, the probability of which is proportional to the logits of the position head and the logits of the LM head. The optimization process is further detailed in Algorithm \ref{alg:beamsearch}. The baselines are implemented by the EvoProtGrad \cite{emami2023plug} package. We perform experiments for 20 times, each comprising 10 optimization rounds.

\begin{table}[tpb]
\centering
\captionsetup{font={small,stretch=0.95}}
\caption{\textbf{Prompts for navigating mutation engineering.}}
\label{tab:fitopt}
\begin{tabular}{ll}
\toprule
Dataset & Prompt                                                               \\
\midrule
AAV     & Increased viability for packaging of a DNA payload for gene therapy. \\
AMIE    & Increase in activity.                                                \\
avGFP   & Leads to enhanced fluorescence at 37 degrees Celsius.                \\
E4B     & Enhances cleavage by caspase-6 and granzyme B.                       \\
LGK     & Increased enzyme activity.                                           \\
UBE2I   & Increased growth rescue rate at high temperature in a yeast strain.  \\
\bottomrule
\end{tabular}
\end{table}

\begin{algorithm}[tpb]
\caption{Multi-round Optimization with Beam Search}\label{alg:beamsearch}
\begin{algorithmic}
\Require Wild-type Sequence $x_{\text{wt}}$, Instruction $t$, Number of Rounds $N$, Number of Candidates $K$
\State $C \gets \{x_{\text{wt}}\}$
\For{Round $=1,2,\cdots,N$}
\For{$x\in C$}
    \State $h\gets f_{\text{PLM}}(x)$
    \State $h\gets h + Decoder(h, T)$\Comment{Add mutational features}
    \State $\text{Score}^{\text{pos}}, \text{Score}^{\text{aa}}\gets f_{\text{pos}}(h), f_{\text{LM}}(h)$\Comment{Calculate the logits two prediction heads}
    \State $\text{Score}(x,i,j)\gets \text{Score}^{\text{pos}}_i+\text{Score}^{\text{aa}}_{i,j},\forall i\neq j$\Comment{The score mutating $i$-th amino acid to $j$}
\EndFor
\State $P(x,i,j)\gets \text{GlobalSoftMax}[\text{Score}(x,i,j)]$\Comment{Probality distribution of sampling mutations}
\State $C\gets \text{Mutate}(x, i, j), (x,i,j)\sim \text{SampleK}(P)$\Comment{Sampling without replacement}
\EndFor
\State \Return $C$
\end{algorithmic}
\end{algorithm}

\subsection{Additional Experiment Results \label{app:result}}

\subsubsection{Experiment Results on Temporal Split} \label{app:temporal-split}

The experimental results for mutation explanation and engineering are shown in Tab. \ref{tab:app-temporal-explanation-result} and Tab. \ref{tab:app-temporal-engineering-result} respectively. We observe that: (1) MutaPLM achieves promising performance on the temporal split and outperforms strong baselines, showcasing its robustness in handling novel mutations. (2) For mutation explanation, the experiment results are similar to those on the \textit{Hard} set of the structural split, and we observe similar over-fitting issues as in structural split that more training steps lead to improved validation loss but performance drops on the test set. This further underscores the significance of improving the generalization capability of mutation explanation models to assist real-world applications. (3) For mutation engineering, the results show little difference with those on the structural split. As discussed in Sec. \ref{exp:eng}, the PLM may have witnessed the protein sequence during pre-training, which mitigates the overfitting problem. 

\begin{table}[tpb]
\centering
\captionsetup{font={small,stretch=0.95}}
\caption{\textbf{Performance evaluation for mutation explanation on temporal split.}}
\setlength\tabcolsep{3pt}
\begin{tabular}{lcccccc}
\toprule
Model & BLEU-2 & BLEU-4 & METEOR & ROUGE-1 & ROUGE-2 & ROUGE-L\\
\midrule
ProLLaMA \cite{lv2024prollama}              & 0.69 & 0.21 & 3.33 & 0.83 & 0.04 & 0.80\\
Galactica-6.7B \cite{taylor2022galactica}   & 3.50 & 1.31 & 5.61 & 7.44 & 0.85 & 6.17 \\
Mol-Instructions \cite{fang2023mol}         & 0.58 & 0.08 & 4.90 & 5.41 & 0.13 & 4.55 \\
\midrule
GPT-4-0613 (5-shot, kNN) \cite{achiam2023gpt}  & 9.30 & 4.25 & 15.08 & 13.92 & 2.29 & 11.84  \\
\midrule
AugmentedESM \cite{hsu2022learning}  & 7.00 & 3.12 & 11.29 & 12.03 & 2.84 & 10.12 \\
Fine-tuned ESM-2 \cite{lin2023evolutionary} & 6.90 & 3.83 & 13.86 & 14.21 & 4.63 & 12.62 \\
\midrule
MutaPLM & \textbf{10.83} & \textbf{6.15} & \textbf{17.84} & \textbf{18.99} & \textbf{6.92} & \textbf{16.51} \\
\bottomrule
\end{tabular}
\label{tab:app-temporal-explanation-result}
\end{table}

\begin{table}[tpb]
\centering
\captionsetup{font={small,stretch=0.95}}
\caption{\textbf{Performance evaluation for mutation engineering on temporal split.}}
\setlength\tabcolsep{4pt}
\begin{tabular}{lccc}
\toprule
Model & Accuracy (\%) &	Recall@50 (\%) \\
\midrule
Random & 4.40 & 0.81 \\
ProtST (ESM-2) \cite{xu2023protst} & 5.11 & - \\
GPT-4-0613 (5-shot, kNN) \cite{achiam2023gpt}  & 12.13 & 6.28 \\
\midrule
ESM-2 \cite{lin2023evolutionary} & 34.76 & 24.02 \\
OntoProtein \cite{zhang2021ontoprotein}  & 37.74 & 28.49 \\
\midrule
Fine-tuned BioMedGPT \cite{luo2023biomedgpt} & 34.57 & 4.09 & \\
Fine-tuned ESM-2 \cite{lin2023evolutionary, gu2021domain} & 55.78 & 44.04 \\
\midrule
MutaPLM & \textbf{58.50} & \textbf{46.05} \\
\bottomrule
\end{tabular}
\label{tab:app-temporal-engineering-result}
\end{table}

\begin{figure}[tpb]
    \centering
    \captionsetup{font={small,stretch=0.95}}
    \includegraphics[width=0.9\linewidth]{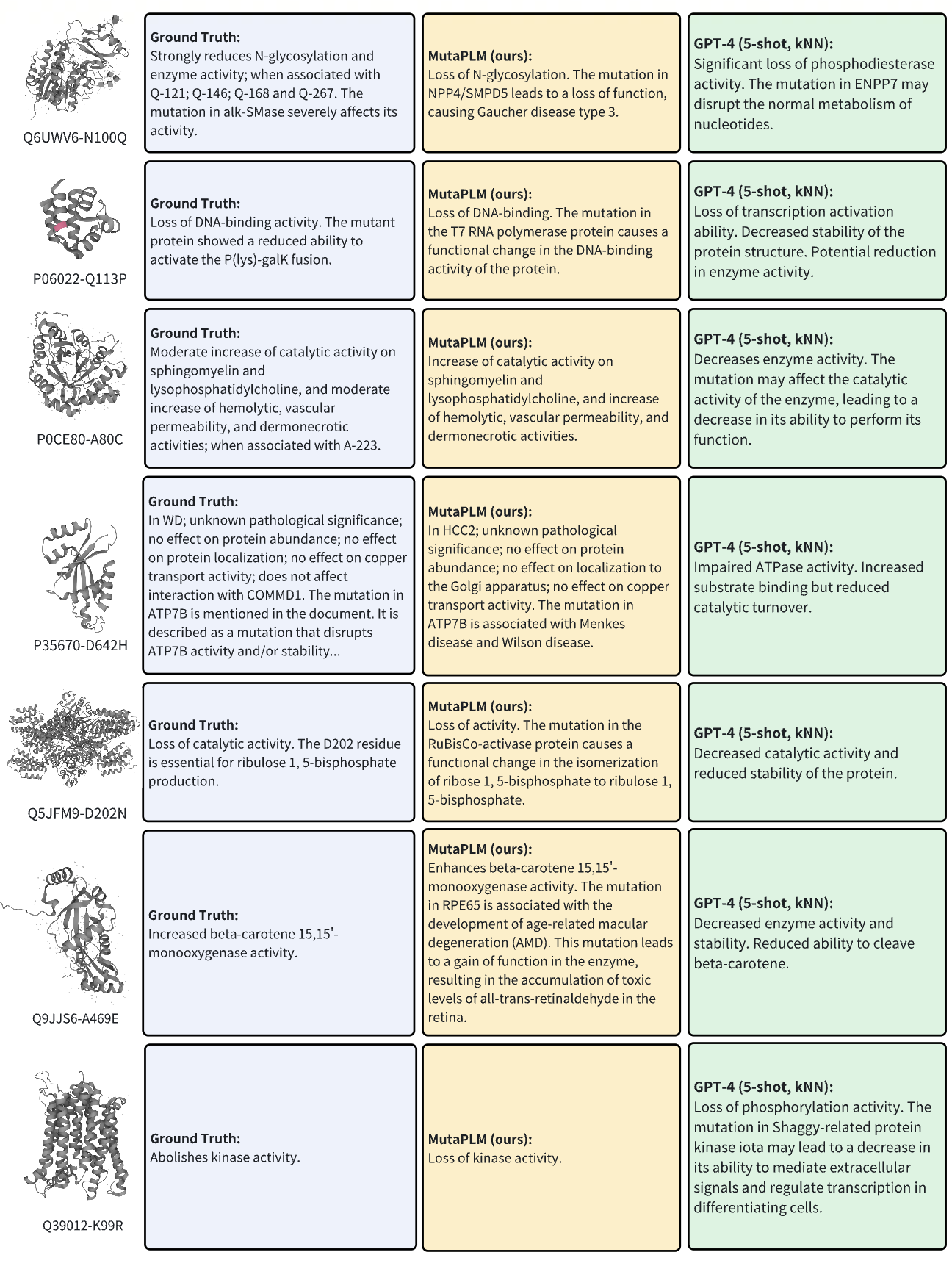}
    \caption{\textbf{More case studies at mutation explanation.} We report the outputs of MutaPLM and GPT-4 (5-shot, kNN).}
    \label{fig:app-more-case}
\end{figure}

\subsubsection{Low-N Fitness Regression}
While MutaPLM is not specifically designed for numeric tasks, we investigate if the learned \textit{Delta} features could benefit fitness regression. We perform experiments on two protein fitness benchmarks, namely Spike-ACE2 \cite{starr2022shifting} and avGFP \cite{sarkisyan2016local}. Spike-ACE2 is a deep mutational scanning dataset that aims to predict the binding strengths between SARS-Cov-2 variants and its receptor ACE2, which is critical for identifying potentially dangerous strains of the virus. The avGFP benchmark aims to predict the fluorescence intensity of GFP variants, which is beneficial for developing biomarkers.

Following prior works \cite{biswas2021low, zhao2024contrastive}, we adopt the low-$N$ setting with 192 randomly sampled training samples and 48 validation samples. We calculate the adaptive fitness by our PLM following \cite{meier2021language} and concatenate it with the \textit{delta} features $z_\Delta$. The result is fed into a trainable 2-layer MLP to predict the fitness scores, and the remaining parameters are kept frozen. We also implement baselines including Ridge Regression, ESM-2 \cite{lin2023evolutionary}, AugmentedESM \cite{hsu2022learning}, Augmented EVmutation \cite{hopf2017mutation}, ConFit \cite{zhao2024contrastive}, and Tranception\_L \cite{notin2022tranception}. All the models are trained for 50 epochs with a batch size of 16 and a learning rate of 0.001 using the MSE loss. We sample different low-$N$ datasets with 5 random seeds and report the results in Tab. \ref{tab:fitness_regress}.

\begin{table}[tpb]
\captionsetup{font={small,stretch=0.95}}
\centering
\caption{{\textbf{Performance evaluation on protein fitness regression benchmarks.}} We perform experiments 5 times with different random seeds and report the Spearman correlation coefficient. The best and second-best results are marked in bold and underlined.}
\label{tab:fitness_regress}
\begin{tabular}{lcc}
\toprule
Model            & Spike-ACE2 & avGFP \\
\midrule
Ridge Regression & 0.335$\pm$0.052 & 0.298$\pm$0.071    \\
ESM-2 \cite{lin2023evolutionary}        & 0.331$\pm$0.041  & 0.554$\pm$0.013    \\
Augmented ESM \cite{hsu2022learning}   & 0.363$\pm$0.021  & 0.497$\pm$0.096    \\
Augmented EVmutation \cite{hopf2017mutation}       & 0.354$\pm$0.044           & 0.512$\pm$0.034    \\
ConFit \cite{zhao2024contrastive} & 0.412$\pm$0.033          & 0.564$\pm$0.035          \\
Tranception\_L \cite{notin2022tranception}      & \textbf{0.488$\pm$0.040}           & \underline{0.594$\pm$0.019}    \\ 
\midrule
MutaPLM   & \underline{0.481$\pm$0.028} & \textbf{0.596$\pm$0.032}   \\
\bottomrule
\end{tabular}
\end{table}

We observe that MutaPLM significantly outperforms baseline models that adopt ESM-2 as the PLM, indicating that the \textit{delta} features have captured mutational knowledge from natural language supervision that benefits fitness regression tasks. While MutaPLM achieves comparable results with Tranception\_L on both benchmarks, it is worth noting that the model adopts a different network architecture specifically designed for fitness regression. Therefore, we speculate that adopting a mutation-oriented PLM instead of ESM-2 may further improve the performance. While fitness regression is not the main focus of our work, we expect future endeavors that jointly harvest discrete textual descriptions and continuous fitness scores. 

\subsubsection{Additional Case Studies} \label{app:detailed-result}
We present more case studies of mutation explanation in Fig. \ref{fig:app-more-case}.

%% file: 7_checklist.tex
\section*{NeurIPS Paper Checklist}

\begin{enumerate}

\item {\bf Claims}
    \item[] Question: Do the main claims made in the abstract and introduction accurately reflect the paper's contributions and scope?
    \item[] Answer: \answerYes{} 
    \item[] Justification: The claims are validated by our experiments in Section \ref{sec:exp}.

\item {\bf Limitations}
    \item[] Question: Does the paper discuss the limitations of the work performed by the authors?
    \item[] Answer: \answerYes{} 
    \item[] Justification: We discuss our limitations in Section \ref{sec:lim}.

\item {\bf Theory Assumptions and Proofs}
    \item[] Question: For each theoretical result, does the paper provide the full set of assumptions and a complete (and correct) proof?
    \item[] Answer: \answerNA{} 
    \item[] Justification: The paper does not include theoretical results.

    \item {\bf Experimental Result Reproducibility}
    \item[] Question: Does the paper fully disclose all the information needed to reproduce the main experimental results of the paper to the extent that it affects the main claims and/or conclusions of the paper (regardless of whether the code and data are provided or not)?
    \item[] Answer: \answerYes{} 
    \item[] Justification: We provide the dataset construction process in Section \ref{sec:dataset} and implementation details in Section \ref{exp:setup}.

\item {\bf Open access to data and code}
    \item[] Question: Does the paper provide open access to the data and code, with sufficient instructions to faithfully reproduce the main experimental results, as described in supplemental material?
    \item[] Answer: \answerYes{} 
    \item[] Justification: We provide our data and code at \url{https://github.com/PharMolix/MutaPLM}.

\item {\bf Experimental Setting/Details}
    \item[] Question: Does the paper specify all the training and test details (e.g., data splits, hyperparameters, how they were chosen, type of optimizer, etc.) necessary to understand the results?
    \item[] Answer: \answerYes{} 
    \item[] Justification: We present implementation details in Section \ref{exp:setup}.

\item {\bf Experiment Statistical Significance}
    \item[] Question: Does the paper report error bars suitably and correctly defined or other appropriate information about the statistical significance of the experiments?
    \item[] Answer: \answerNo{} 
    \item[] Justification: While we report error bars for protein fitness optimization, the majority of experiments do not include error bars because it would be too computationally expensive.

\item {\bf Experiments Compute Resources}
    \item[] Question: For each experiment, does the paper provide sufficient information on the computer resources (type of compute workers, memory, time of execution) needed to reproduce the experiments?
    \item[] Answer: \answerYes{} 
    \item[] Justification: The information is provided in Section \ref{exp:setup}.
    
\item {\bf Code Of Ethics}
    \item[] Question: Does the research conducted in the paper conform, in every respect, with the NeurIPS Code of Ethics \url{https://neurips.cc/public/EthicsGuidelines}?
    \item[] Answer: \answerYes{} 
    \item[] Justification: N/A.

\item {\bf Broader Impacts}
    \item[] Question: Does the paper discuss both potential positive societal impacts and negative societal impacts of the work performed?
    \item[] Answer: \answerYes{} 
    \item[] Justification: We discuss broader impacts in Section \ref{sec:lim}.
    
\item {\bf Safeguards}
    \item[] Question: Does the paper describe safeguards that have been put in place for responsible release of data or models that have a high risk for misuse (e.g., pretrained language models, image generators, or scraped datasets)?
    \item[] Answer: \answerYes{} 
    \item[] Justification: We clarify safeguards in Section \ref{sec:lim}.

\item {\bf Licenses for existing assets}
    \item[] Question: Are the creators or original owners of assets (e.g., code, data, models), used in the paper, properly credited and are the license and terms of use explicitly mentioned and properly respected?
    \item[] Answer: \answerYes{} 
    \item[] Justification: The original papers of assets are cited.

\item {\bf New Assets}
    \item[] Question: Are new assets introduced in the paper well documented and is the documentation provided alongside the assets?
    \item[] Answer: \answerYes{} 
    \item[] Justification: The assets and documentation are at \url{https://github.com/PharMolix/MutaPLM}.

\item {\bf Crowdsourcing and Research with Human Subjects}
    \item[] Question: For crowdsourcing experiments and research with human subjects, does the paper include the full text of instructions given to participants and screenshots, if applicable, as well as details about compensation (if any)? 
    \item[] Answer: \answerYes{} 
    \item[] Justification: The instructions to human participants are displayed in Appendix \ref{app: prompt_eva}.

\item {\bf Institutional Review Board (IRB) Approvals or Equivalent for Research with Human Subjects}
    \item[] Question: Does the paper describe potential risks incurred by study participants, whether such risks were disclosed to the subjects, and whether Institutional Review Board (IRB) approvals (or an equivalent approval/review based on the requirements of your country or institution) were obtained?
    \item[] Answer: \answerNA{} 
    \item[] Justification: The paper does not involve crowdsourcing nor research with human subjects.

\end{enumerate}

%% file: neurips_2024.bbl
\begin{thebibliography}{99}
\providecommand{\natexlab}[1]{#1}
\providecommand{\url}[1]{\texttt{#1}}
\expandafter\ifx\csname urlstyle\endcsname\relax
  \providecommand{\doi}[1]{doi: #1}\else
  \providecommand{\doi}{doi: \begingroup \urlstyle{rm}\Url}\fi

\bibitem[P{\'a}l et~al.(2006)P{\'a}l, Papp, and Lercher]{pal2006integrated}
Csaba P{\'a}l, Bal{\'a}zs Papp, and Martin~J Lercher.
\newblock An integrated view of protein evolution.
\newblock \emph{Nature reviews genetics}, 7\penalty0 (5):\penalty0 337--348, 2006.

\bibitem[Soskine and Tawfik(2010)]{soskine2010mutational}
Misha Soskine and Dan~S Tawfik.
\newblock Mutational effects and the evolution of new protein functions.
\newblock \emph{Nature Reviews Genetics}, 11\penalty0 (8):\penalty0 572--582, 2010.

\bibitem[Reva et~al.(2011)Reva, Antipin, and Sander]{reva2011predicting}
Boris Reva, Yevgeniy Antipin, and Chris Sander.
\newblock Predicting the functional impact of protein mutations: application to cancer genomics.
\newblock \emph{Nucleic acids research}, 39\penalty0 (17):\penalty0 e118--e118, 2011.

\bibitem[Harvey et~al.(2021)Harvey, Carabelli, Jackson, Gupta, Thomson, Harrison, Ludden, Reeve, Rambaut, Peacock, et~al.]{harvey2021sars}
William~T Harvey, Alessandro~M Carabelli, Ben Jackson, Ravindra~K Gupta, Emma~C Thomson, Ewan~M Harrison, Catherine Ludden, Richard Reeve, Andrew Rambaut, Sharon~J Peacock, et~al.
\newblock Sars-cov-2 variants, spike mutations and immune escape.
\newblock \emph{Nature Reviews Microbiology}, 19\penalty0 (7):\penalty0 409--424, 2021.

\bibitem[Hu et~al.(2022)Hu, Peng, Cao, Wu, Chen, Wang, Tang, and Huang]{hu2022increased}
Jie Hu, Pai Peng, Xiaoxia Cao, Kang Wu, Juan Chen, Kai Wang, Ni~Tang, and Ai-long Huang.
\newblock Increased immune escape of the new sars-cov-2 variant of concern omicron.
\newblock \emph{Cellular \& Molecular Immunology}, 19\penalty0 (2):\penalty0 293--295, 2022.

\bibitem[Thomas et~al.(1995)Thomas, Qu, and Pedersen]{thomas1995defective}
Philip~J Thomas, Bao-He Qu, and Peter~L Pedersen.
\newblock Defective protein folding as a basis of human disease.
\newblock \emph{Trends in biochemical sciences}, 20\penalty0 (11):\penalty0 456--459, 1995.

\bibitem[Dobson(2001)]{dobson2001structural}
Christopher~M Dobson.
\newblock The structural basis of protein folding and its links with human disease.
\newblock \emph{Philosophical Transactions of the Royal Society of London. Series B: Biological Sciences}, 356\penalty0 (1406):\penalty0 133--145, 2001.

\bibitem[Turner(2009)]{turner2009directed}
Nicholas~J Turner.
\newblock Directed evolution drives the next generation of biocatalysts.
\newblock \emph{Nature chemical biology}, 5\penalty0 (8):\penalty0 567--573, 2009.

\bibitem[Arnold(2018)]{arnold2018directed}
Frances~H Arnold.
\newblock Directed evolution: bringing new chemistry to life.
\newblock \emph{Angewandte Chemie (International Ed. in English)}, 57\penalty0 (16):\penalty0 4143, 2018.

\bibitem[Arnold and Volkov(1999)]{arnold1999directed}
Frances~H Arnold and Alexander~A Volkov.
\newblock Directed evolution of biocatalysts.
\newblock \emph{Current opinion in chemical biology}, 3\penalty0 (1):\penalty0 54--59, 1999.

\bibitem[Sawano and Miyawaki(2000)]{sawano2000directed}
Asako Sawano and Atsushi Miyawaki.
\newblock Directed evolution of green fluorescent protein by a new versatile pcr strategy for site-directed and semi-random mutagenesis.
\newblock \emph{Nucleic acids research}, 28\penalty0 (16):\penalty0 e78--e78, 2000.

\bibitem[Boder et~al.(2000)Boder, Midelfort, and Wittrup]{boder2000directed}
Eric~T Boder, Katarina~S Midelfort, and K~Dane Wittrup.
\newblock Directed evolution of antibody fragments with monovalent femtomolar antigen-binding affinity.
\newblock \emph{Proceedings of the National Academy of Sciences}, 97\penalty0 (20):\penalty0 10701--10705, 2000.

\bibitem[Sun et~al.(2017)Sun, Zhou, Lai, and Pei]{sun2017sequence}
Tanlin Sun, Bo~Zhou, Luhua Lai, and Jianfeng Pei.
\newblock Sequence-based prediction of protein protein interaction using a deep-learning algorithm.
\newblock \emph{BMC bioinformatics}, 18:\penalty0 1--8, 2017.

\bibitem[Gao et~al.(2020)Gao, Mahajan, Sulam, and Gray]{gao2020deep}
Wenhao Gao, Sai~Pooja Mahajan, Jeremias Sulam, and Jeffrey~J Gray.
\newblock Deep learning in protein structural modeling and design.
\newblock \emph{Patterns}, 1\penalty0 (9), 2020.

\bibitem[Bepler and Berger(2021)]{bepler2021learning}
Tristan Bepler and Bonnie Berger.
\newblock Learning the protein language: Evolution, structure, and function.
\newblock \emph{Cell systems}, 12\penalty0 (6):\penalty0 654--669, 2021.

\bibitem[Madani et~al.(2020)Madani, McCann, Naik, Keskar, Anand, Eguchi, Huang, and Socher]{madani2020progen}
Ali Madani, Bryan McCann, Nikhil Naik, Nitish~Shirish Keskar, Namrata Anand, Raphael~R Eguchi, Po-Ssu Huang, and Richard Socher.
\newblock Progen: Language modeling for protein generation.
\newblock \emph{arXiv preprint arXiv:2004.03497}, 2020.

\bibitem[Elnaggar et~al.(2021)Elnaggar, Heinzinger, Dallago, Rehawi, Wang, Jones, Gibbs, Feher, Angerer, Steinegger, et~al.]{elnaggar2021prottrans}
Ahmed Elnaggar, Michael Heinzinger, Christian Dallago, Ghalia Rehawi, Yu~Wang, Llion Jones, Tom Gibbs, Tamas Feher, Christoph Angerer, Martin Steinegger, et~al.
\newblock Prottrans: Toward understanding the language of life through self-supervised learning.
\newblock \emph{IEEE transactions on pattern analysis and machine intelligence}, 44\penalty0 (10):\penalty0 7112--7127, 2021.

\bibitem[Ferruz et~al.(2022)Ferruz, Schmidt, and H{\"o}cker]{ferruz2022protgpt2}
Noelia Ferruz, Steffen Schmidt, and Birte H{\"o}cker.
\newblock Protgpt2 is a deep unsupervised language model for protein design.
\newblock \emph{Nature communications}, 13\penalty0 (1):\penalty0 4348, 2022.

\bibitem[Lin et~al.(2023)Lin, Akin, Rao, Hie, Zhu, Lu, Smetanin, Verkuil, Kabeli, Shmueli, et~al.]{lin2023evolutionary}
Zeming Lin, Halil Akin, Roshan Rao, Brian Hie, Zhongkai Zhu, Wenting Lu, Nikita Smetanin, Robert Verkuil, Ori Kabeli, Yaniv Shmueli, et~al.
\newblock Evolutionary-scale prediction of atomic-level protein structure with a language model.
\newblock \emph{Science}, 379\penalty0 (6637):\penalty0 1123--1130, 2023.

\bibitem[Truong~Jr and Bepler(2024)]{truong2024poet}
Timothy Truong~Jr and Tristan Bepler.
\newblock Poet: A generative model of protein families as sequences-of-sequences.
\newblock \emph{Advances in Neural Information Processing Systems}, 36, 2024.

\bibitem[Liu et~al.(2021)Liu, Zhang, Hou, Mian, Wang, Zhang, and Tang]{liu2021self}
Xiao Liu, Fanjin Zhang, Zhenyu Hou, Li~Mian, Zhaoyu Wang, Jing Zhang, and Jie Tang.
\newblock Self-supervised learning: Generative or contrastive.
\newblock \emph{IEEE transactions on knowledge and data engineering}, 35\penalty0 (1):\penalty0 857--876, 2021.

\bibitem[Suzek et~al.(2015)Suzek, Wang, Huang, McGarvey, Wu, and Consortium]{suzek2015uniref}
Baris~E Suzek, Yuqi Wang, Hongzhan Huang, Peter~B McGarvey, Cathy~H Wu, and UniProt Consortium.
\newblock Uniref clusters: a comprehensive and scalable alternative for improving sequence similarity searches.
\newblock \emph{Bioinformatics}, 31\penalty0 (6):\penalty0 926--932, 2015.

\bibitem[Steinegger and S{\"o}ding(2018)]{steinegger2018clustering}
Martin Steinegger and Johannes S{\"o}ding.
\newblock Clustering huge protein sequence sets in linear time.
\newblock \emph{Nature communications}, 9\penalty0 (1):\penalty0 2542, 2018.

\bibitem[Weissenow et~al.(2022)Weissenow, Heinzinger, and Rost]{weissenow2022protein}
Konstantin Weissenow, Michael Heinzinger, and Burkhard Rost.
\newblock Protein language-model embeddings for fast, accurate, and alignment-free protein structure prediction.
\newblock \emph{Structure}, 30\penalty0 (8):\penalty0 1169--1177, 2022.

\bibitem[Ferruz and H{\"o}cker(2022)]{ferruz2022controllable}
Noelia Ferruz and Birte H{\"o}cker.
\newblock Controllable protein design with language models.
\newblock \emph{Nature Machine Intelligence}, 4\penalty0 (6):\penalty0 521--532, 2022.

\bibitem[Meier et~al.(2021)Meier, Rao, Verkuil, Liu, Sercu, and Rives]{meier2021language}
Joshua Meier, Roshan Rao, Robert Verkuil, Jason Liu, Tom Sercu, and Alex Rives.
\newblock Language models enable zero-shot prediction of the effects of mutations on protein function.
\newblock \emph{Advances in neural information processing systems}, 34:\penalty0 29287--29303, 2021.

\bibitem[Hsu et~al.(2022)Hsu, Nisonoff, Fannjiang, and Listgarten]{hsu2022learning}
Chloe Hsu, Hunter Nisonoff, Clara Fannjiang, and Jennifer Listgarten.
\newblock Learning protein fitness models from evolutionary and assay-labeled data.
\newblock \emph{Nature biotechnology}, 40\penalty0 (7):\penalty0 1114--1122, 2022.

\bibitem[Zhao et~al.(2024)Zhao, Zhang, and Luo]{zhao2024contrastive}
Junming Zhao, Chao Zhang, and Yunan Luo.
\newblock Contrastive fitness learning: Reprogramming protein language models for low-n learning of protein fitness landscape.
\newblock In \emph{International Conference on Research in Computational Molecular Biology}, pages 470--474. Springer, 2024.

\bibitem[Vaswani et~al.(2017)Vaswani, Shazeer, Parmar, Uszkoreit, Jones, Gomez, Kaiser, and Polosukhin]{vaswani2017attention}
Ashish Vaswani, Noam Shazeer, Niki Parmar, Jakob Uszkoreit, Llion Jones, Aidan~N Gomez, {\L}ukasz Kaiser, and Illia Polosukhin.
\newblock Attention is all you need.
\newblock \emph{Advances in neural information processing systems}, 30, 2017.

\bibitem[Wei et~al.(2022)Wei, Wang, Schuurmans, Bosma, Xia, Chi, Le, Zhou, et~al.]{wei2022chain}
Jason Wei, Xuezhi Wang, Dale Schuurmans, Maarten Bosma, Fei Xia, Ed~Chi, Quoc~V Le, Denny Zhou, et~al.
\newblock Chain-of-thought prompting elicits reasoning in large language models.
\newblock \emph{Advances in neural information processing systems}, 35:\penalty0 24824--24837, 2022.

\bibitem[Touvron et~al.(2023)Touvron, Martin, Stone, Albert, Almahairi, Babaei, Bashlykov, Batra, Bhargava, Bhosale, et~al.]{touvron2023llama}
Hugo Touvron, Louis Martin, Kevin Stone, Peter Albert, Amjad Almahairi, Yasmine Babaei, Nikolay Bashlykov, Soumya Batra, Prajjwal Bhargava, Shruti Bhosale, et~al.
\newblock Llama 2: Open foundation and fine-tuned chat models.
\newblock \emph{arXiv preprint arXiv:2307.09288}, 2023.

\bibitem[Jiang et~al.(2023)Jiang, Sablayrolles, Mensch, Bamford, Chaplot, Casas, Bressand, Lengyel, Lample, Saulnier, et~al.]{jiang2023mistral}
Albert~Q Jiang, Alexandre Sablayrolles, Arthur Mensch, Chris Bamford, Devendra~Singh Chaplot, Diego de~las Casas, Florian Bressand, Gianna Lengyel, Guillaume Lample, Lucile Saulnier, et~al.
\newblock Mistral 7b.
\newblock \emph{arXiv preprint arXiv:2310.06825}, 2023.

\bibitem[Achiam et~al.(2023)Achiam, Adler, Agarwal, Ahmad, Akkaya, Aleman, Almeida, Altenschmidt, Altman, Anadkat, et~al.]{achiam2023gpt}
Josh Achiam, Steven Adler, Sandhini Agarwal, Lama Ahmad, Ilge Akkaya, Florencia~Leoni Aleman, Diogo Almeida, Janko Altenschmidt, Sam Altman, Shyamal Anadkat, et~al.
\newblock Gpt-4 technical report.
\newblock \emph{arXiv preprint arXiv:2303.08774}, 2023.

\bibitem[Bi et~al.(2024)Bi, Chen, Chen, Chen, Dai, Deng, Ding, Dong, Du, Fu, et~al.]{bi2024deepseek}
Xiao Bi, Deli Chen, Guanting Chen, Shanhuang Chen, Damai Dai, Chengqi Deng, Honghui Ding, Kai Dong, Qiushi Du, Zhe Fu, et~al.
\newblock Deepseek llm: Scaling open-source language models with longtermism.
\newblock \emph{arXiv preprint arXiv:2401.02954}, 2024.

\bibitem[Brandes et~al.(2022)Brandes, Ofer, Peleg, Rappoport, and Linial]{brandes2022proteinbert}
Nadav Brandes, Dan Ofer, Yam Peleg, Nadav Rappoport, and Michal Linial.
\newblock Proteinbert: a universal deep-learning model of protein sequence and function.
\newblock \emph{Bioinformatics}, 38\penalty0 (8):\penalty0 2102--2110, 2022.

\bibitem[Rives et~al.(2021)Rives, Meier, Sercu, Goyal, Lin, Liu, Guo, Ott, Zitnick, Ma, et~al.]{rives2021biological}
Alexander Rives, Joshua Meier, Tom Sercu, Siddharth Goyal, Zeming Lin, Jason Liu, Demi Guo, Myle Ott, C~Lawrence Zitnick, Jerry Ma, et~al.
\newblock Biological structure and function emerge from scaling unsupervised learning to 250 million protein sequences.
\newblock \emph{Proceedings of the National Academy of Sciences}, 118\penalty0 (15):\penalty0 e2016239118, 2021.

\bibitem[Hayes et~al.(2024)Hayes, Rao, Akin, Sofroniew, Oktay, Lin, Verkuil, Tran, Deaton, Wiggert, et~al.]{hayes2024simulating}
Tomas Hayes, Roshan Rao, Halil Akin, Nicholas~J Sofroniew, Deniz Oktay, Zeming Lin, Robert Verkuil, Vincent~Q Tran, Jonathan Deaton, Marius Wiggert, et~al.
\newblock Simulating 500 million years of evolution with a language model.
\newblock \emph{bioRxiv}, pages 2024--07, 2024.

\bibitem[Devlin et~al.(2019)Devlin, Chang, Lee, and Toutanova]{devlin2019bert}
Jacob Devlin, Ming-Wei Chang, Kenton Lee, and Kristina Toutanova.
\newblock Bert: Pre-training of deep bidirectional transformers for language understanding.
\newblock In \emph{Proceedings of the 2019 Conference of the North American Chapter of the Association for Computational Linguistics: Human Language Technologies, Volume 1 (Long and Short Papers)}, pages 4171--4186, 2019.

\bibitem[Radford et~al.(2018)Radford, Narasimhan, Salimans, Sutskever, et~al.]{radford2018improving}
Alec Radford, Karthik Narasimhan, Tim Salimans, Ilya Sutskever, et~al.
\newblock Improving language understanding by generative pre-training.
\newblock 2018.

\bibitem[Unsal et~al.(2022)Unsal, Atas, Albayrak, Turhan, Acar, and Do{\u{g}}an]{unsal2022learning}
Serbulent Unsal, Heval Atas, Muammer Albayrak, Kemal Turhan, Aybar~C Acar, and Tunca Do{\u{g}}an.
\newblock Learning functional properties of proteins with language models.
\newblock \emph{Nature Machine Intelligence}, 4\penalty0 (3):\penalty0 227--245, 2022.

\bibitem[Hu et~al.(2024)Hu, Li, Rao, Thafar, and Arif]{hu2024improving}
Jun Hu, Zhe Li, Bing Rao, Maha~A Thafar, and Muhammad Arif.
\newblock Improving protein-protein interaction prediction using protein language model and protein network features.
\newblock \emph{Analytical Biochemistry}, page 115550, 2024.

\bibitem[Xu et~al.(2023)Xu, Yuan, Miret, and Tang]{xu2023protst}
Minghao Xu, Xinyu Yuan, Santiago Miret, and Jian Tang.
\newblock Protst: Multi-modality learning of protein sequences and biomedical texts.
\newblock In \emph{International Conference on Machine Learning}, pages 38749--38767. PMLR, 2023.

\bibitem[Luo et~al.(2024)Luo, Liu, Yang, Huang, Hong, Zhang, Wu, and Nie]{luo2024toward}
Yizhen Luo, Xing~Yi Liu, Kai Yang, Kui Huang, Massimo Hong, Jiahuan Zhang, Yushuai Wu, and Zaiqing Nie.
\newblock Toward unified ai drug discovery with multimodal knowledge.
\newblock \emph{Health Data Science}, 4:\penalty0 0113, 2024.

\bibitem[Zhuo et~al.(2024)Zhuo, Chi, Xu, Huang, Zheng, He, Mao, and Zhang]{zhuo2024protllm}
Le~Zhuo, Zewen Chi, Minghao Xu, Heyan Huang, Heqi Zheng, Conghui He, Xian-Ling Mao, and Wentao Zhang.
\newblock Protllm: An interleaved protein-language llm with protein-as-word pre-training.
\newblock \emph{arXiv preprint arXiv:2403.07920}, 2024.

\bibitem[Lv et~al.(2024)Lv, Lin, Li, Liu, Cui, Chen, Yuan, and Tian]{lv2024prollama}
Liuzhenghao Lv, Zongying Lin, Hao Li, Yuyang Liu, Jiaxi Cui, Calvin Yu-Chian Chen, Li~Yuan, and Yonghong Tian.
\newblock Prollama: A protein large language model for multi-task protein language processing.
\newblock \emph{arXiv preprint arXiv:2402.16445}, 2024.

\bibitem[Yin et~al.(2024)Yin, Zhou, Zhu, Lin, Wu, Wu, Xu, Hsieh, Hou, Chen, et~al.]{yin2024multi}
Mingze Yin, Hanjing Zhou, Yiheng Zhu, Miao Lin, Yixuan Wu, Jialu Wu, Hongxia Xu, Chang-Yu Hsieh, Tingjun Hou, Jintai Chen, et~al.
\newblock Multi-modal clip-informed protein editing.
\newblock \emph{bioRxiv}, pages 2024--07, 2024.

\bibitem[Romero and Arnold(2009)]{romero2009exploring}
Philip~A Romero and Frances~H Arnold.
\newblock Exploring protein fitness landscapes by directed evolution.
\newblock \emph{Nature reviews Molecular cell biology}, 10\penalty0 (12):\penalty0 866--876, 2009.

\bibitem[Hopf et~al.(2017)Hopf, Ingraham, Poelwijk, Sch{\"a}rfe, Springer, Sander, and Marks]{hopf2017mutation}
Thomas~A Hopf, John~B Ingraham, Frank~J Poelwijk, Charlotta~PI Sch{\"a}rfe, Michael Springer, Chris Sander, and Debora~S Marks.
\newblock Mutation effects predicted from sequence co-variation.
\newblock \emph{Nature biotechnology}, 35\penalty0 (2):\penalty0 128--135, 2017.

\bibitem[Laine et~al.(2019)Laine, Karami, and Carbone]{laine2019gemme}
Elodie Laine, Yasaman Karami, and Alessandra Carbone.
\newblock Gemme: a simple and fast global epistatic model predicting mutational effects.
\newblock \emph{Molecular biology and evolution}, 36\penalty0 (11):\penalty0 2604--2619, 2019.

\bibitem[Jeanmougin et~al.(1998)Jeanmougin, Thompson, Gouy, Higgins, and Gibson]{jeanmougin1998multiple}
Francois Jeanmougin, Julie~D Thompson, Manolo Gouy, Desmond~G Higgins, and Toby~J Gibson.
\newblock Multiple sequence alignment with clustal x.
\newblock \emph{Trends in biochemical sciences}, 23\penalty0 (10):\penalty0 403--405, 1998.

\bibitem[Dauparas et~al.(2022)Dauparas, Anishchenko, Bennett, Bai, Ragotte, Milles, Wicky, Courbet, de~Haas, Bethel, et~al.]{dauparas2022robust}
Justas Dauparas, Ivan Anishchenko, Nathaniel Bennett, Hua Bai, Robert~J Ragotte, Lukas~F Milles, Basile~IM Wicky, Alexis Courbet, Rob~J de~Haas, Neville Bethel, et~al.
\newblock Robust deep learning--based protein sequence design using proteinmpnn.
\newblock \emph{Science}, 378\penalty0 (6615):\penalty0 49--56, 2022.

\bibitem[Rao et~al.(2021)Rao, Liu, Verkuil, Meier, Canny, Abbeel, Sercu, and Rives]{rao2021msa}
Roshan~M Rao, Jason Liu, Robert Verkuil, Joshua Meier, John Canny, Pieter Abbeel, Tom Sercu, and Alexander Rives.
\newblock Msa transformer.
\newblock In \emph{International Conference on Machine Learning}, pages 8844--8856. PMLR, 2021.

\bibitem[Notin et~al.(2022{\natexlab{a}})Notin, Van~Niekerk, Kollasch, Ritter, Gal, and Marks]{notin2022trancepteve}
Pascal Notin, Lood Van~Niekerk, Aaron~W Kollasch, Daniel Ritter, Yarin Gal, and Debora~Susan Marks.
\newblock Trancepteve: Combining family-specific and family-agnostic models of protein sequences for improved fitness prediction.
\newblock In \emph{NeurIPS 2022 Workshop on Learning Meaningful Representations of Life}, 2022{\natexlab{a}}.

\bibitem[Fowler and Fields(2014)]{fowler2014deep}
Douglas~M Fowler and Stanley Fields.
\newblock Deep mutational scanning: a new style of protein science.
\newblock \emph{Nature methods}, 11\penalty0 (8):\penalty0 801--807, 2014.

\bibitem[Landrum and Kattman(2018)]{landrum2018clinvar}
Melissa~J Landrum and Brandi~L Kattman.
\newblock Clinvar at five years: Delivering on the promise.
\newblock \emph{Human mutation}, 39\penalty0 (11):\penalty0 1623--1630, 2018.

\bibitem[Brookes et~al.(2019)Brookes, Park, and Listgarten]{brookes2019conditioning}
David Brookes, Hahnbeom Park, and Jennifer Listgarten.
\newblock Conditioning by adaptive sampling for robust design.
\newblock In \emph{International conference on machine learning}, pages 773--782. PMLR, 2019.

\bibitem[Stanton et~al.(2022)Stanton, Maddox, Gruver, Maffettone, Delaney, Greenside, and Wilson]{stanton2022accelerating}
Samuel Stanton, Wesley Maddox, Nate Gruver, Phillip Maffettone, Emily Delaney, Peyton Greenside, and Andrew~Gordon Wilson.
\newblock Accelerating bayesian optimization for biological sequence design with denoising autoencoders.
\newblock In \emph{International Conference on Machine Learning}, pages 20459--20478. PMLR, 2022.

\bibitem[Gruver et~al.(2024)Gruver, Stanton, Frey, Rudner, Hotzel, Lafrance-Vanasse, Rajpal, Cho, and Wilson]{gruver2024protein}
Nate Gruver, Samuel Stanton, Nathan Frey, Tim~GJ Rudner, Isidro Hotzel, Julien Lafrance-Vanasse, Arvind Rajpal, Kyunghyun Cho, and Andrew~G Wilson.
\newblock Protein design with guided discrete diffusion.
\newblock \emph{Advances in Neural Information Processing Systems}, 36, 2024.

\bibitem[Sinai et~al.(2020)Sinai, Wang, Whatley, Slocum, Locane, and Kelsic]{sinai2020adalead}
Sam Sinai, Richard Wang, Alexander Whatley, Stewart Slocum, Elina Locane, and Eric~D Kelsic.
\newblock Adalead: A simple and robust adaptive greedy search algorithm for sequence design.
\newblock \emph{arXiv preprint arXiv:2010.02141}, 2020.

\bibitem[Angermueller et~al.(2019)Angermueller, Dohan, Belanger, Deshpande, Murphy, and Colwell]{angermueller2019model}
Christof Angermueller, David Dohan, David Belanger, Ramya Deshpande, Kevin Murphy, and Lucy Colwell.
\newblock Model-based reinforcement learning for biological sequence design.
\newblock In \emph{International conference on learning representations}, 2019.

\bibitem[Kirjner et~al.(2023)Kirjner, Yim, Samusevich, Jaakkola, Barzilay, and Fiete]{kirjner2023optimizing}
Andrew Kirjner, Jason Yim, Raman Samusevich, Tommi~S Jaakkola, Regina Barzilay, and Ila~R Fiete.
\newblock Optimizing protein fitness using gibbs sampling with graph-based smoothing.
\newblock In \emph{ICML 2023 Workshop: Sampling and Optimization in Discrete Space}, 2023.

\bibitem[Luo et~al.(2023{\natexlab{a}})Luo, Zhang, Fan, Yang, Wu, Qiao, and Nie]{luo2023biomedgpt}
Yizhen Luo, Jiahuan Zhang, Siqi Fan, Kai Yang, Yushuai Wu, Mu~Qiao, and Zaiqing Nie.
\newblock Biomedgpt: Open multimodal generative pre-trained transformer for biomedicine.
\newblock \emph{arXiv preprint arXiv:2308.09442}, 2023{\natexlab{a}}.

\bibitem[Ke et~al.(2022)Ke, Shao, Lin, Konishi, Kim, and Liu]{ke2022continual}
Zixuan Ke, Yijia Shao, Haowei Lin, Tatsuya Konishi, Gyuhak Kim, and Bing Liu.
\newblock Continual pre-training of language models.
\newblock In \emph{The Eleventh International Conference on Learning Representations}, 2022.

\bibitem[Li et~al.(2022)Li, Li, Xiong, and Hoi]{li2022blip}
Junnan Li, Dongxu Li, Caiming Xiong, and Steven Hoi.
\newblock Blip: Bootstrapping language-image pre-training for unified vision-language understanding and generation.
\newblock In \emph{International conference on machine learning}, pages 12888--12900. PMLR, 2022.

\bibitem[Li et~al.(2023)Li, Li, Savarese, and Hoi]{li2023blip}
Junnan Li, Dongxu Li, Silvio Savarese, and Steven Hoi.
\newblock Blip-2: Bootstrapping language-image pre-training with frozen image encoders and large language models.
\newblock In \emph{International conference on machine learning}, pages 19730--19742. PMLR, 2023.

\bibitem[Zheng et~al.(2023)Zheng, Deng, Xue, Zhou, Ye, and Gu]{zheng2023structure}
Zaixiang Zheng, Yifan Deng, Dongyu Xue, Yi~Zhou, Fei Ye, and Quanquan Gu.
\newblock Structure-informed language models are protein designers.
\newblock In \emph{International Conference on Machine Learning}, pages 42317--42338. PMLR, 2023.

\bibitem[Fang et~al.(2023)Fang, Liang, Zhang, Liu, Huang, Chen, Fan, and Chen]{fang2023mol}
Yin Fang, Xiaozhuan Liang, Ningyu Zhang, Kangwei Liu, Rui Huang, Zhuo Chen, Xiaohui Fan, and Huajun Chen.
\newblock Mol-instructions-a large-scale biomolecular instruction dataset for large language models.
\newblock In \emph{The Twelfth International Conference on Learning Representations}, 2023.

\bibitem[Ghazvininejad et~al.(2019)Ghazvininejad, Levy, Liu, and Zettlemoyer]{ghazvininejad2019mask}
Marjan Ghazvininejad, Omer Levy, Yinhan Liu, and Luke Zettlemoyer.
\newblock Mask-predict: Parallel decoding of conditional masked language models.
\newblock \emph{arXiv preprint arXiv:1904.09324}, 2019.

\bibitem[Boutet et~al.(2016)Boutet, Lieberherr, Tognolli, Schneider, Bansal, Bridge, Poux, Bougueleret, and Xenarios]{boutet2016uniprotkb}
Emmanuel Boutet, Damien Lieberherr, Michael Tognolli, Michel Schneider, Parit Bansal, Alan~J Bridge, Sylvain Poux, Lydie Bougueleret, and Ioannis Xenarios.
\newblock Uniprotkb/swiss-prot, the manually annotated section of the uniprot knowledgebase: how to use the entry view.
\newblock \emph{Plant bioinformatics: methods and protocols}, pages 23--54, 2016.

\bibitem[Canese and Weis(2013)]{canese2013pubmed}
Kathi Canese and Sarah Weis.
\newblock Pubmed: the bibliographic database.
\newblock \emph{The NCBI handbook}, 2\penalty0 (1), 2013.

\bibitem[Steinegger and S{\"o}ding(2017)]{steinegger2017mmseqs2}
Martin Steinegger and Johannes S{\"o}ding.
\newblock Mmseqs2 enables sensitive protein sequence searching for the analysis of massive data sets.
\newblock \emph{Nature biotechnology}, 35\penalty0 (11):\penalty0 1026--1028, 2017.

\bibitem[Riesselman et~al.(2018)Riesselman, Ingraham, and Marks]{riesselman2018deep}
Adam~J Riesselman, John~B Ingraham, and Debora~S Marks.
\newblock Deep generative models of genetic variation capture the effects of mutations.
\newblock \emph{Nature methods}, 15\penalty0 (10):\penalty0 816--822, 2018.

\bibitem[Notin et~al.(2024)Notin, Kollasch, Ritter, Van~Niekerk, Paul, Spinner, Rollins, Shaw, Orenbuch, Weitzman, et~al.]{notin2024proteingym}
Pascal Notin, Aaron Kollasch, Daniel Ritter, Lood Van~Niekerk, Steffanie Paul, Han Spinner, Nathan Rollins, Ada Shaw, Rose Orenbuch, Ruben Weitzman, et~al.
\newblock Proteingym: large-scale benchmarks for protein fitness prediction and design.
\newblock \emph{Advances in Neural Information Processing Systems}, 36, 2024.

\bibitem[Taylor et~al.(2022)Taylor, Kardas, Cucurull, Scialom, Hartshorn, Saravia, Poulton, Kerkez, and Stojnic]{taylor2022galactica}
Ross Taylor, Marcin Kardas, Guillem Cucurull, Thomas Scialom, Anthony Hartshorn, Elvis Saravia, Andrew Poulton, Viktor Kerkez, and Robert Stojnic.
\newblock Galactica: A large language model for science.
\newblock \emph{arXiv preprint arXiv:2211.09085}, 2022.

\bibitem[Zhang et~al.(2021)Zhang, Bi, Liang, Cheng, Hong, Deng, Zhang, Lian, and Chen]{zhang2021ontoprotein}
Ningyu Zhang, Zhen Bi, Xiaozhuan Liang, Siyuan Cheng, Haosen Hong, Shumin Deng, Qiang Zhang, Jiazhang Lian, and Huajun Chen.
\newblock Ontoprotein: Protein pretraining with gene ontology embedding.
\newblock In \emph{International Conference on Learning Representations}, 2021.

\bibitem[Luo et~al.(2023{\natexlab{b}})Luo, Yang, Meng, Li, Zhou, and Zhang]{luo2023empirical}
Yun Luo, Zhen Yang, Fandong Meng, Yafu Li, Jie Zhou, and Yue Zhang.
\newblock An empirical study of catastrophic forgetting in large language models during continual fine-tuning.
\newblock \emph{arXiv preprint arXiv:2308.08747}, 2023{\natexlab{b}}.

\bibitem[Hu et~al.(2021)Hu, Wallis, Allen-Zhu, Li, Wang, Wang, Chen, et~al.]{hu2021lora}
Edward~J Hu, Phillip Wallis, Zeyuan Allen-Zhu, Yuanzhi Li, Shean Wang, Lu~Wang, Weizhu Chen, et~al.
\newblock Lora: Low-rank adaptation of large language models.
\newblock In \emph{International Conference on Learning Representations}, 2021.

\bibitem[Loshchilov and Hutter(2018)]{loshchilov2018decoupled}
Ilya Loshchilov and Frank Hutter.
\newblock Decoupled weight decay regularization.
\newblock In \emph{International Conference on Learning Representations}, 2018.

\bibitem[Dong et~al.(2022)Dong, Li, Dai, Zheng, Wu, Chang, Sun, Xu, and Sui]{dong2022survey}
Qingxiu Dong, Lei Li, Damai Dai, Ce~Zheng, Zhiyong Wu, Baobao Chang, Xu~Sun, Jingjing Xu, and Zhifang Sui.
\newblock A survey on in-context learning.
\newblock \emph{arXiv preprint arXiv:2301.00234}, 2022.

\bibitem[AI4Science and Quantum(2023)]{ai4science2023impact}
Microsoft~Research AI4Science and Microsoft~Azure Quantum.
\newblock The impact of large language models on scientific discovery: a preliminary study using gpt-4.
\newblock \emph{arXiv preprint arXiv:2311.07361}, 2023.

\bibitem[Papineni et~al.(2002)Papineni, Roukos, Ward, and Zhu]{papineni2002bleu}
Kishore Papineni, Salim Roukos, Todd Ward, and Wei-Jing Zhu.
\newblock Bleu: a method for automatic evaluation of machine translation.
\newblock In \emph{Proceedings of the 40th annual meeting of the Association for Computational Linguistics}, pages 311--318, 2002.

\bibitem[Lin(2004)]{lin2004rouge}
Chin-Yew Lin.
\newblock Rouge: A package for automatic evaluation of summaries.
\newblock In \emph{Text summarization branches out}, pages 74--81, 2004.

\bibitem[Gu et~al.(2021)Gu, Tinn, Cheng, Lucas, Usuyama, Liu, Naumann, Gao, and Poon]{gu2021domain}
Yu~Gu, Robert Tinn, Hao Cheng, Michael Lucas, Naoto Usuyama, Xiaodong Liu, Tristan Naumann, Jianfeng Gao, and Hoifung Poon.
\newblock Domain-specific language model pretraining for biomedical natural language processing.
\newblock \emph{ACM Transactions on Computing for Healthcare (HEALTH)}, 3\penalty0 (1):\penalty0 1--23, 2021.

\bibitem[Ren et~al.(2021)Ren, Xiao, Chang, Huang, Li, Gupta, Chen, and Wang]{ren2021survey}
Pengzhen Ren, Yun Xiao, Xiaojun Chang, Po-Yao Huang, Zhihui Li, Brij~B Gupta, Xiaojiang Chen, and Xin Wang.
\newblock A survey of deep active learning.
\newblock \emph{ACM computing surveys (CSUR)}, 54\penalty0 (9):\penalty0 1--40, 2021.

\bibitem[Freitag and Al-Onaizan(2017)]{freitag2017beam}
Markus Freitag and Yaser Al-Onaizan.
\newblock Beam search strategies for neural machine translation.
\newblock In \emph{Proceedings of the First Workshop on Neural Machine Translation}, pages 56--60, 2017.

\bibitem[Ren et~al.(2022)Ren, Li, Ding, Zhou, Ma, and Peng]{ren2022proximal}
Zhizhou Ren, Jiahan Li, Fan Ding, Yuan Zhou, Jianzhu Ma, and Jian Peng.
\newblock Proximal exploration for model-guided protein sequence design.
\newblock In \emph{International Conference on Machine Learning}, pages 18520--18536. PMLR, 2022.

\bibitem[Emami et~al.(2023)Emami, Perreault, Law, Biagioni, and John]{emami2023plug}
Patrick Emami, Aidan Perreault, Jeffrey Law, David Biagioni, and Peter~St John.
\newblock Plug \& play directed evolution of proteins with gradient-based discrete mcmc.
\newblock \emph{Machine Learning: Science and Technology}, 4\penalty0 (2):\penalty0 025014, 2023.

\bibitem[Abramson et~al.(2024)Abramson, Adler, Dunger, Evans, Green, Pritzel, Ronneberger, Willmore, Ballard, Bambrick, et~al.]{abramson2024accurate}
Josh Abramson, Jonas Adler, Jack Dunger, Richard Evans, Tim Green, Alexander Pritzel, Olaf Ronneberger, Lindsay Willmore, Andrew~J Ballard, Joshua Bambrick, et~al.
\newblock Accurate structure prediction of biomolecular interactions with alphafold 3.
\newblock \emph{Nature}, pages 1--3, 2024.

\bibitem[Zhu et~al.(2023)Zhu, Chen, Shen, Li, and Elhoseiny]{zhu2023minigpt}
Deyao Zhu, Jun Chen, Xiaoqian Shen, Xiang Li, and Mohamed Elhoseiny.
\newblock Minigpt-4: Enhancing vision-language understanding with advanced large language models.
\newblock \emph{arXiv preprint arXiv:2304.10592}, 2023.

\bibitem[Radford et~al.(2021)Radford, Kim, Hallacy, Ramesh, Goh, Agarwal, Sastry, Askell, Mishkin, Clark, et~al.]{radford2021learning}
Alec Radford, Jong~Wook Kim, Chris Hallacy, Aditya Ramesh, Gabriel Goh, Sandhini Agarwal, Girish Sastry, Amanda Askell, Pamela Mishkin, Jack Clark, et~al.
\newblock Learning transferable visual models from natural language supervision.
\newblock In \emph{International conference on machine learning}, pages 8748--8763. PMLR, 2021.

\bibitem[Bryant et~al.(2021)Bryant, Bashir, Sinai, Jain, Ogden, Riley, Church, Colwell, and Kelsic]{bryant2021deep}
Drew~H Bryant, Ali Bashir, Sam Sinai, Nina~K Jain, Pierce~J Ogden, Patrick~F Riley, George~M Church, Lucy~J Colwell, and Eric~D Kelsic.
\newblock Deep diversification of an aav capsid protein by machine learning.
\newblock \emph{Nature Biotechnology}, 39\penalty0 (6):\penalty0 691--696, 2021.

\bibitem[Wrenbeck et~al.(2017)Wrenbeck, Azouz, and Whitehead]{wrenbeck2017single}
Emily~E Wrenbeck, Laura~R Azouz, and Timothy~A Whitehead.
\newblock Single-mutation fitness landscapes for an enzyme on multiple substrates reveal specificity is globally encoded.
\newblock \emph{Nature communications}, 8\penalty0 (1):\penalty0 15695, 2017.

\bibitem[Sarkisyan et~al.(2016)Sarkisyan, Bolotin, Meer, Usmanova, Mishin, Sharonov, Ivankov, Bozhanova, Baranov, Soylemez, et~al.]{sarkisyan2016local}
Karen~S Sarkisyan, Dmitry~A Bolotin, Margarita~V Meer, Dinara~R Usmanova, Alexander~S Mishin, George~V Sharonov, Dmitry~N Ivankov, Nina~G Bozhanova, Mikhail~S Baranov, Onuralp Soylemez, et~al.
\newblock Local fitness landscape of the green fluorescent protein.
\newblock \emph{Nature}, 533\penalty0 (7603):\penalty0 397--401, 2016.

\bibitem[Starita et~al.(2013)Starita, Pruneda, Lo, Fowler, Kim, Hiatt, Shendure, Brzovic, Fields, and Klevit]{starita2013activity}
Lea~M Starita, Jonathan~N Pruneda, Russell~S Lo, Douglas~M Fowler, Helen~J Kim, Joseph~B Hiatt, Jay Shendure, Peter~S Brzovic, Stanley Fields, and Rachel~E Klevit.
\newblock Activity-enhancing mutations in an e3 ubiquitin ligase identified by high-throughput mutagenesis.
\newblock \emph{Proceedings of the National Academy of Sciences}, 110\penalty0 (14):\penalty0 E1263--E1272, 2013.

\bibitem[Klesmith et~al.(2015)Klesmith, Bacik, Michalczyk, and Whitehead]{klesmith2015comprehensive}
Justin~R Klesmith, John-Paul Bacik, Ryszard Michalczyk, and Timothy~A Whitehead.
\newblock Comprehensive sequence-flux mapping of a levoglucosan utilization pathway in e. coli.
\newblock \emph{ACS synthetic biology}, 4\penalty0 (11):\penalty0 1235--1243, 2015.

\bibitem[Weile et~al.(2017)Weile, Sun, Cote, Knapp, Verby, Mellor, Wu, Pons, Wong, van Lieshout, et~al.]{weile2017framework}
Jochen Weile, Song Sun, Atina~G Cote, Jennifer Knapp, Marta Verby, Joseph~C Mellor, Yingzhou Wu, Carles Pons, Cassandra Wong, Natascha van Lieshout, et~al.
\newblock A framework for exhaustively mapping functional missense variants.
\newblock \emph{Molecular systems biology}, 13\penalty0 (12):\penalty0 957, 2017.

\bibitem[Starr et~al.(2022)Starr, Greaney, Hannon, Loes, Hauser, Dillen, Ferri, Farrell, Dadonaite, McCallum, et~al.]{starr2022shifting}
Tyler~N Starr, Allison~J Greaney, William~W Hannon, Andrea~N Loes, Kevin Hauser, Josh~R Dillen, Elena Ferri, Ariana~Ghez Farrell, Bernadeta Dadonaite, Matthew McCallum, et~al.
\newblock Shifting mutational constraints in the sars-cov-2 receptor-binding domain during viral evolution.
\newblock \emph{Science}, 377\penalty0 (6604):\penalty0 420--424, 2022.

\bibitem[Biswas et~al.(2021)Biswas, Khimulya, Alley, Esvelt, and Church]{biswas2021low}
Surojit Biswas, Grigory Khimulya, Ethan~C Alley, Kevin~M Esvelt, and George~M Church.
\newblock Low-n protein engineering with data-efficient deep learning.
\newblock \emph{Nature methods}, 18\penalty0 (4):\penalty0 389--396, 2021.

\bibitem[Notin et~al.(2022{\natexlab{b}})Notin, Dias, Frazer, Marchena-Hurtado, Gomez, Marks, and Gal]{notin2022tranception}
Pascal Notin, Mafalda Dias, Jonathan Frazer, Javier Marchena-Hurtado, Aidan~N Gomez, Debora Marks, and Yarin Gal.
\newblock Tranception: protein fitness prediction with autoregressive transformers and inference-time retrieval.
\newblock In \emph{International Conference on Machine Learning}, pages 16990--17017. PMLR, 2022{\natexlab{b}}.

\end{thebibliography}
